\documentclass{article}

\usepackage{arxiv}

\usepackage{cite}
\usepackage{graphicx}
\usepackage{booktabs}
\usepackage{array}
\usepackage{multirow}
\usepackage[dvipsnames]{xcolor}
\definecolor{flavescent}{rgb}{0.97, 0.91, 0.56}

\usepackage{subcaption}
\usepackage{url}
\usepackage{tabularx}
\newcolumntype{M}[1]{>{\centering\arraybackslash}m{#1}}

\newcommand{\ra}[1]{\renewcommand{\arraystretch}{#1}}

\usepackage{amssymb}
\usepackage{xcolor,colortbl}

\definecolor{Gray}{gray}{0.85}
\newcolumntype{a}{>{\columncolor{Gray}}c}

\title{Evaluating Single Image Dehazing Methods Under Realistic Sunlight Haze}

\author{
  Zahra Anvari \\
  Department of Computer Science and Engineering\\
  University of Texas Arlington\\
  Arlington, TX \\
  \texttt{zahra.anvari@mavs.uta.edu} \\
   \And
 Vassilis Athitsos \\
  Department of Computer Science and Engineering\\
  University of Texas Arlington\\
  Arlington, TX \\
  \texttt{athitsos@uta.edu} \\
}

\begin{document}
\maketitle

\begin{abstract}
Haze can degrade the visibility and the image quality drastically, thus degrading the performance of computer vision tasks such as object detection. Single image dehazing is a challenging and ill-posed problem, despite being widely studied. 
Most existing methods assume that haze has a uniform/homogeneous distribution and haze can have a single color,~\textit{i.e.} grayish white color similar to smoke, while in reality haze can be distributed non-uniformly with different patterns and colors. 
In this paper, we focus on haze created by sunlight as it is one of the most prevalent type of haze in the wild. Sunlight can generate non-uniformly distributed haze with drastic density changes due to sun rays and also a spectrum of haze color due to sunlight color changes during the day. 
This presents a new challenge to image dehazing methods. For these methods to be practical, this problem needs to be addressed. 
To quantify the challenges and assess the performance of these methods, we present a sunlight haze benchmark dataset, Sun-Haze, containing 107 hazy images with different types of haze created by sunlight having a variety of intensity and color. We evaluate a representative set of state-of-the-art image dehazing methods on this benchmark dataset in terms of standard metrics such as PSNR, SSIM, CIEDE2000, PI and NIQE. This uncovers the limitation of the current methods, and questions their underlying assumptions as well as their practicality.
 \end{abstract}

% keywords can be removed

\section{Introduction}
Haze is an atmospheric phenomenon that can cause poor visibility, low contrast, and image quality degradation. Haze can decrease the performance of different computer vision tasks, such as object detection~\cite{liu2016ssd,redmon2016you}, semantic segmentation~\cite{long2015fully}, face detection, clustering, and dataset creation~\cite{yang2016wider,anvari2019pipeline,lin2018deep,lin2017proximity,schroff2015facenet},~\textit{etc}. Therefore, haze removal has drawn a great deal of attention over the past decade. 

\begin{figure}[h]
    \centering
    \begin{subfigure}[b]{0.25\textwidth}
        \centering
\includegraphics[width=0.9\textwidth]{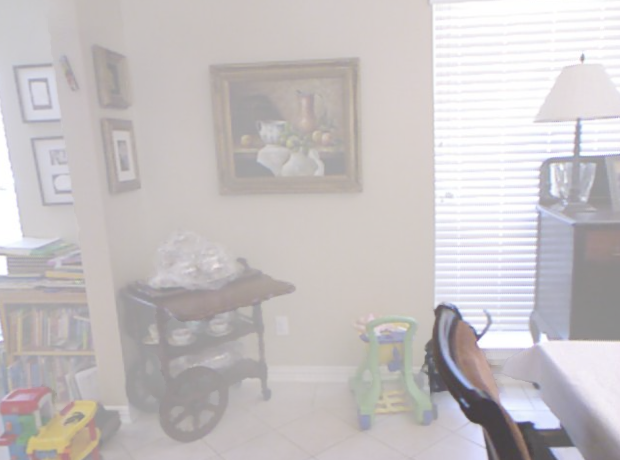}
        \label{fig:hazy}
        \caption{SOTS indoor}
    \end{subfigure}
    \hspace*{-0.1in}%
    \begin{subfigure}[b]{0.25\textwidth}  
        \centering 
        \includegraphics[width=0.9\textwidth]{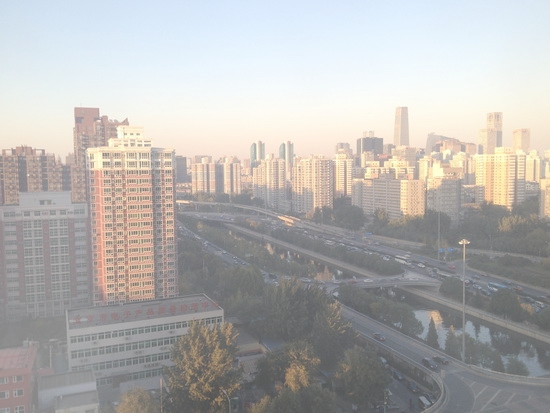}
        \label{fig:aodnet}
         \caption{SOTS outdoor}
    \end{subfigure}
    \hspace*{-0.1in}%
    \begin{subfigure}[b]{0.25\textwidth}   
        \centering 
        \includegraphics[width=0.9\textwidth]{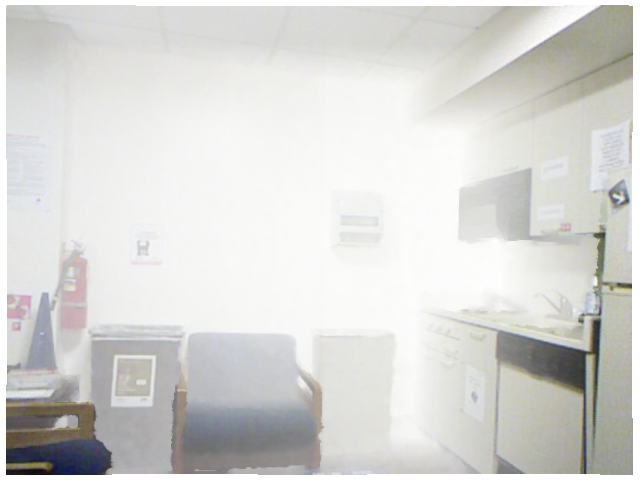}
        \label{fig:epdn}
        \caption{NYU}
    \end{subfigure}
    \hspace*{-0.1in}%
    \begin{subfigure}[b]{0.25\textwidth}   
        \centering 
        \includegraphics[width=0.9\textwidth]{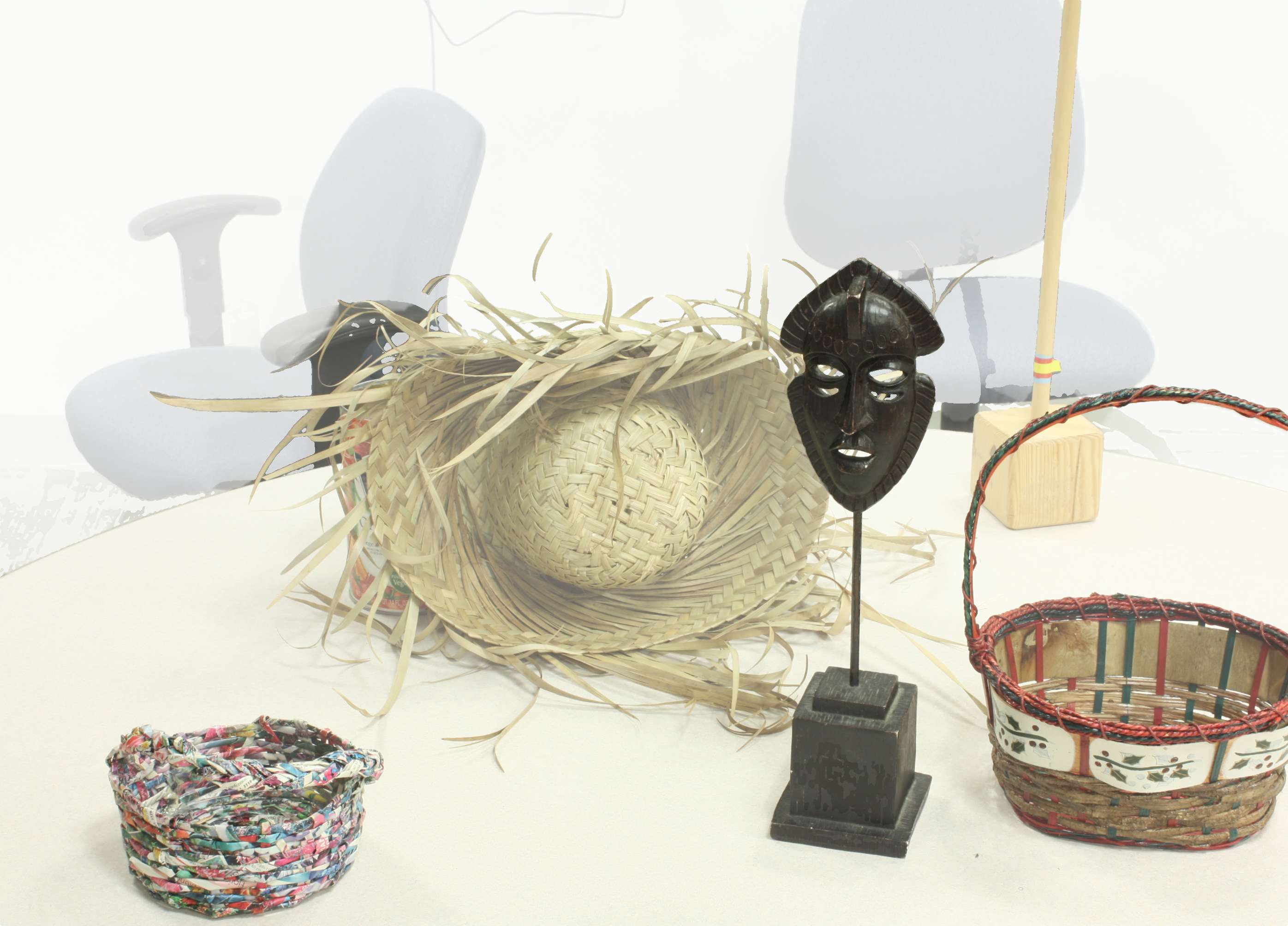}
        \label{fig:ours1}
        \caption{Middlebury}
    \end{subfigure}
    \caption{Sample hazy images of the datasets widely used to test image dehazing methods, SOTS test dataset~\cite{li2018benchmarking}, NYU dataset~\cite{ancuti2016d}, and Middlebury~\cite{ancuti2016d}.}
    {\small } 
    \label{fig:others}
\end{figure}

However, haze removal is still a challenging and ill-posed problem and most existing methods make assumptions that do not simply hold in reality. For example, most existing methods and datasets assume that i) haze has a uniform and homogeneous distribution in the entire image, and ii) haze can only have a single color,~\textit{i.e.} grayish white similar to the color of smoke or pollution. While in reality haze density can change non-homogeneously throughout an image and it can vary in pattern and color. Figure~\ref{fig:others} shows a few sample images of different haze datasets that are widely used to test image dehazing methods. As you can see, haze is monochromatic and homogeneous in all these images. These datasets are created synthetically and do not look realistic, which may limit the practicality of dehazing methods.

In this paper, we focus on haze created by sunlight which present a unique challenge to dehazing methods. The reason we focus on sunlight is that haze created by sunlight is one of the most prevalent type of haze for the outdoor and indoor settings, and yet it has not received enough attention. To the best of our knowledge, our work is the first work that focuses on sunlight haze. 

Haze created by sunlight has multiple unique features: i) it can drastically vary in between sun rays throughout an image, ii) it can corrupt some parts of an image more than another, meaning haze density drastically varies, iii) it can have an spectrum of colors, due to the sunlight color changes during the day, and iv) it has a unique gradually diminishing pattern.

\begin{figure}[t]
    \centering
    \begin{subfigure}[b]{0.33\textwidth}
        \centering
\includegraphics[width=0.8\textwidth]{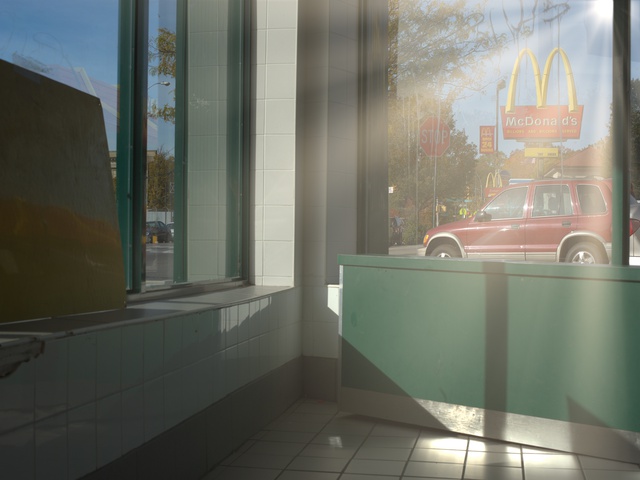}
        \label{fig:hazy}
    \end{subfigure}
    \hspace*{-0.2in}%
    \begin{subfigure}[b]{0.33\textwidth}  
        \centering 
        \includegraphics[width=0.8\textwidth]{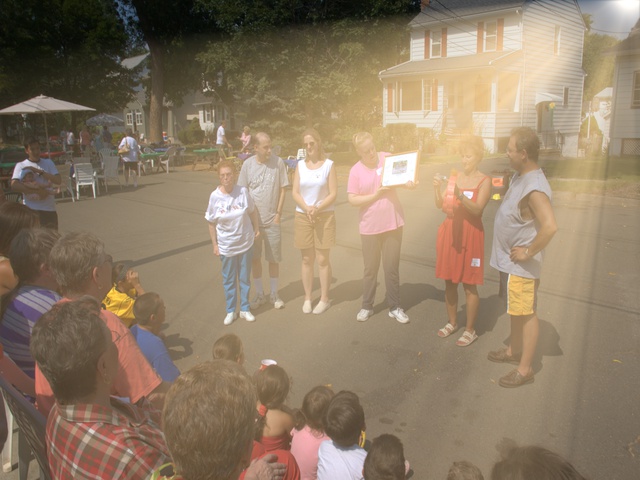}
        \label{fig:aodnet}
    \end{subfigure}
    \hspace*{-0.2in}%
    \begin{subfigure}[b]{0.33\textwidth}   
        \centering 
        \includegraphics[width=0.8\textwidth]{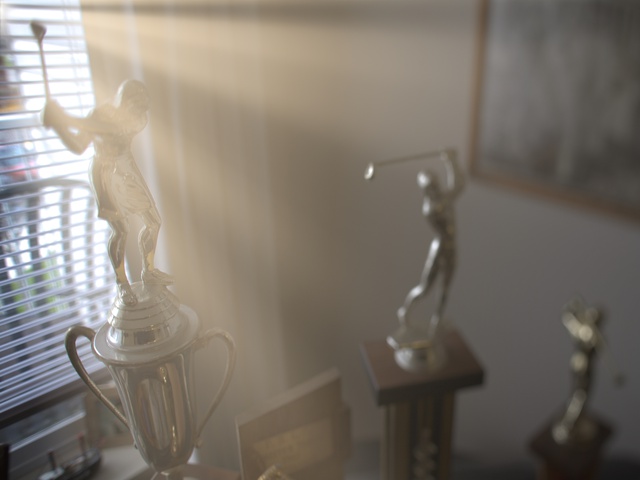}
        \label{fig:epdn}
    \end{subfigure}
    \hspace*{0in}%
    \begin{subfigure}[b]{0.33\textwidth}   
        \centering 
        \includegraphics[width=0.8\textwidth]{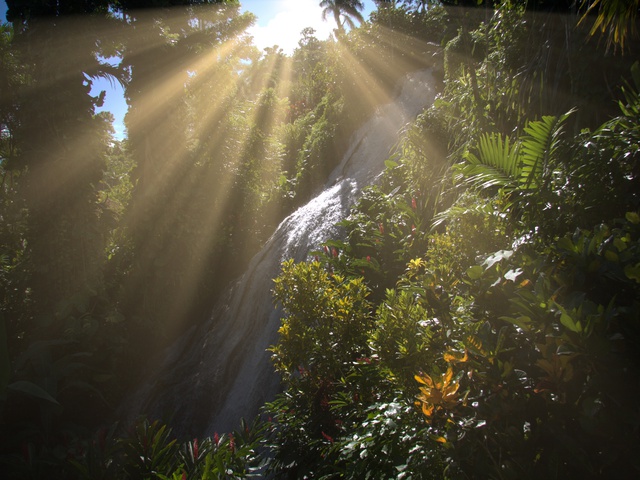}
        \label{fig:ours1}
    \end{subfigure}
    \hspace*{-0.2in}%
    \begin{subfigure}[b]{0.33\textwidth}
    \centering
        \includegraphics[width=0.8\textwidth]{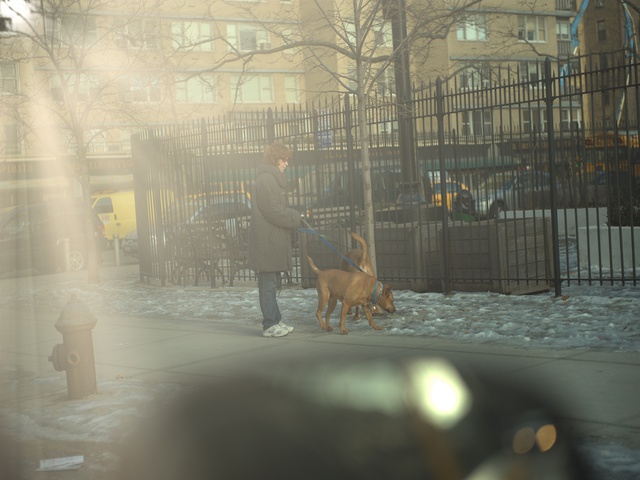}
        \label{fig:hazy}
    \end{subfigure}
    \hspace*{-0.2in}%
    \begin{subfigure}[b]{0.33\textwidth}  
        \centering 
        \includegraphics[width=0.8\textwidth]{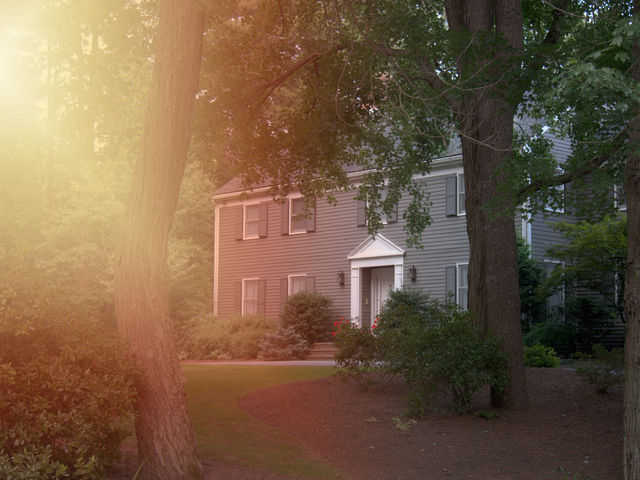}
        \label{fig:aodnet}
    \end{subfigure}
    
    \caption{Sample hazy images of Sun-Haze}
    {\small } 
    \label{fig:sunhaze}
\end{figure}

To quantify the challenges and assess the performance of state-of-the-art dehazing methods, we present a sunlight haze benchmark dataset, Sun-Haze, containing 107 hazy images with different haze density, coverage, and color, caused by sunlight. Figure~\ref{fig:sunhaze} presents some sample hazy images of Sun-Haze. We describe our dataset in the next section in details.

Since the ground truth/haze-free image of a hazy image can be a variety of clean images, for instance images with different contrast or lighting, having only a single image as the ground truth might not be a fair representative and it lacks flexibility and practicality. Thus we build our dataset on top of MIT-Adobe FiveK dataset~\cite{bychkovsky2011learning} which includes images retouched by five experts. The retouched images are clean images that we can employ as ground truth. Therefore, our dataset contains six ground truth images per hazy image, including the original one before adding haze. This provides us with the opportunity to compare existing methods more widely, fairly and more importantly in a more practical way.

Our evaluation of the current state-of-the-art methods shows that there is no clear winner and all these dehazing methods suffer to generalize well to the haze created by sunlight, specially when dealing with haze with a different color.

In summary, this paper presents the following contributions:
\begin{itemize}
    \item We present a sunlight haze benchmark dataset, Sun-Haze, that contains 107 hazy images caused by sunlight, along with six ground truth images (five retouched by five experts and one original image before being retouched) per hazy image.
    \item We perform an extensive analysis to evaluate current state-of-the-art dehazing methods over Sun-Haze in terms of both reference-based and no-reference-based metrics.
    \item We show that existing dehazing methods can not generalize well when there is sunlight haze, in particular when we have sunlight color changes.
\end{itemize}

The rest of this paper is organized as follows. Section 2 explains our dataset, Sun-Haze. Section 3 describes a representative set of state-of-the-art dehazing methods to evaluate on our Sun-Haze dataset. Section 4 presents our evaluation results and discussion.

\section{Sun-Haze dataset}
In this section, we describe how we created our hazy dataset, called Sun-Haze. This dataset is built on top of MIT-Adobe FiveK dataset~\cite{bychkovsky2011learning}. MIT-Adobe FiveK dataset contains 5,000 photos taken by photographers with SLR cameras. 
These photos are all in RAW format, meaning that all the information recorded by the camera sensors,~\emph{i.e.} metadata, is preserved. 
These photos are captured from different scenes, subjects, and during various lighting conditions. These photographs are retouched to obtain a visually pleasing renditions by five photography experts using~\textit{Adobe Lightroom}~\cite{lightroom}.

%Figure~\ref{fig:experts} shows the images retouched by five experts on the input image. As depicted, different retouchers worked on various set of parameters to achieve visually pleasing images, for instance exposure, color density, brightness, etc.   

To create our own dataset, we carefully selected a subset of 107 images from MIT-Adobe FiveK dataset, and added sunlight haze to them. We selected images that would create a realistic image after adding the sunlight haze. For example, we did not select night time images, or indoor images with no windows. 

To add sunlight haze and mimic the real sunlight haze effect, we utilized~\emph{Adobe Photoshop}~\cite{photoshop} and~\textit{Luminar 4}~\cite{luminar4} which are photo editing applications. They enable us to add realistic sunlight haze. To produce realistic sunlight haze effect, we carefully used different parameters in Luminar 4, such as sunlight length, number of sunlight rays, intensity, penetration, and warmth. Increasing intensity would create a thicker and more dense haze effect. Increasing penetration would expand the sunlight haze effect to a broader region of the image. Increasing sunlight warmth creates a golden yellow type of haze, which enables us to create realistic sunlight color changes during the day. We also added sunlight haze from different angles to further diversify our dataset. To create sunset/sunrise haze effect, we used Adobe Photoshop and professionally added a gradient sunlight haze effect.

Sun-Haze dataset includes 107 outdoor and indoor images with sunlight haze professionally added. It also includes the original image (before retouch) as well as five retouched images (retouched by five experts) per hazy image, that serve as the ground truth images for each hazy image. Therefore each hazy image in our dataset has six ground truth images, enabling us to evaluate dehazing methods more widely and in a more practical way (since the ground truth of a hazy image could be a variety of clean/haze-free images). We will make our dataset public, and we hope that this dataset can help other researchers to test dehazing methods in a more practical and realistic way.

\section{Dehazing Methods}\label{existing}
In this section, we describe a representative of existing dehazing methods from the earliest ones to the current state-of-the-art methods. We can categorize the dehazing methods into the following categories (Note that some of the methods might fall into multiple categories):

\begin{itemize}
    \item \textbf{Prior-based:} Prior-based methods also known as prior information-based methods are mainly based on the parameter estimation of atmospheric scattering model by utilizing the priors, such as dark channel priors~\cite{he2010single}, color attenuation prior~\cite{zhang2011atmospheric}, haze-line prior~\cite{berman2016non, berman2017air}. The physical scattering model consists of the transmission map and the atmospheric light, and it is formulated as follows:
    \begin{equation}\label{eq_phy1}
    I(x) = J(x)t(x) + A(1 - t(x))
    \end{equation}
    where $I(x)$ is the hazy image, $J(x)$ is the haze-free image or the scene radiance, $t(x)$ is the medium transmission map, and $A$ is the global atmospheric light on each $x$ pixel coordinates.
    
    \item \textbf{Learning-based:} On the other hand, some methods utilize the deep convolutional neural networks to estimate the transmission map indirectly~\cite{cai2016dehazenet, ren2016single}. Some work employ deep convolutional neural networks to jointly estimate the parameters of the physical scattering model,~\emph{i.e.} atmospheric light and the transmission map~\cite{yang2018towards, zhu2018dehazegan}.  
    \item \textbf{Paired/Unpaired Supervision:} Paired single image dehazing methods need the haze-free/ground truth of each hazy image for training~\cite{qu2019enhanced, li2017aod}, while unpaired dehazing methods do not require the haze-free pair of the hazy images~\cite{anvari2020dehaze, engin2018cycle}. 
    
    \item \textbf{Adversarial-based:} Some image dehazing methods utilize generative adversarial networks for image dehazing and learn transmission map and atmospheric light simultaneously in the generators. Some data-driven methods use adversarial training to solve the dehazing problem without using priors. Some recently proposed work use image-to-image translation techniques to tackle the image dehazing problem through adversarial training~\cite{anvari2020dehaze,qu2019enhanced, engin2018cycle}. Xitong~\emph{et. al} proposed a joint model that learns to perform physical-model based disentanglement by adversarial training~\cite{yang2018towards}.

\end{itemize}

Table~\ref{methods} represents a description of the methods that we evaluated. As you can observe, we selected a variety of methods from different categories. Next, we will describe these methods in more details.

\begin{table}
\ra{1.2}
\caption{Description of the evaluated existing methods.}
\label{methods}
\centering
\setlength{\tabcolsep}{6pt}
\resizebox{1.0\textwidth}{!}{\begin{tabular}{lcccc}
\toprule
 Method & Paired vs. Unpaired & Prior-based & Learning-based & Adversarial-based\\\toprule 
DCP~\cite{he2010single} & NA & \checkmark &  &  \\
MSCNN~\cite{ren2016single} & Paired & \checkmark & \checkmark  & \\
DehazeNet~\cite{cai2016dehazenet} & Paired  & \checkmark & \checkmark &\\
AODNet~\cite{li2017aod} & Paired & \checkmark & \checkmark &     \\
EPDN~\cite{qu2019enhanced} & Paired & & \checkmark  & \checkmark\\
Dehaze-GLCGAN~\cite{anvari2020dehaze} & Unpaired  &  & \checkmark  & \checkmark \\
CycleDehaze~\cite{engin2018cycle} & Unpaired & & \checkmark  & \checkmark\\
\hline
\end{tabular}}
\end{table}

\subsection{DCP}
Early dehazing methods are mostly prior-based methods, DCP~\cite{he2010single} is one of those Prior-based dehazing methods~\cite{he2010single, zhu2015fast} which estimates the transmission map by investigating the dark channel prior. DCP utilizes dark channel prior to more reliably calculate the transmission matrix. With dark channel prior, the thickness of haze is estimated and removed by the atmospheric scattering model. 

Moreover, this method is proposed based on experimental statistics of experiments on haze-free images, which shows at least one color channel has some pixels with very low intensities in most of non-haze patches. However, DCP has poor performance on dehazing the sky images and is computationally intensive.

\subsection{MSCNN}
The performance of prior-based image dehazing methods is limited by hand-designed features, such as the dark channel, color disparity and maximum contrast, with complex fusion schemes. Unlike the previous methods, MSCNN~\cite{ren2016single} is a learning-based dehazing method. The authors proposed a multi-scale deep neural network for single-image dehazing by learning the mapping between hazy images and their corresponding transmission maps.

This method contains two sub-networks called coarse-scale and fine-scale, to estimate the transmission map. The coarse-scale network estimates the transmission map based on the entire image. The results are further improved locally by the fine-scale network.

%The authors  synthesized a dataset comprised of hazy images and the corresponding transmission maps based on the NYU Depth dataset.

\subsection{DehazeNet}
DehazeNet~\cite{cai2016dehazenet} takes advantage of both priors and the power of convolutional neural networks. DehazeNet proposed an end-to-end system for medium transmission estimation. It takes a hazy image as input, and outputs its medium transmission map that is later used to recover a haze-free image via atmospheric scattering model. In addition, DehazeNet proposed a CNN-based deep network, which its layers are specially designed to embody the established priors in image dehazing. Authors also proposed a nonlinear activation function called Bilateral Rectified Linear Unit (BReLU), to improve the quality of recovered haze-free image.

In short, DehazeNet modified the classic CNN model by adding feature extraction and non-linear regression layers. These modifications distinguish DehazeNet from other CNN-based models.

\subsection{AOD-Net}
AOD-Net~\cite{li2017aod} is an end-to-end dehazing  network which is based on estimating the transmission map through reformulating the atmospheric scattering model. Instead of estimating the transmission matrix and the atmospheric light separately, AODNet directly generates the clean image through a light-weight CNN. AOD-Net can be easily embedded with Faster R-CNN~\cite{ren2015faster} and improve the object detection performance on hazy images with a large margin.

\subsection{EPDN}
EPDN~\cite{qu2019enhanced} is a recently proposed GAN-based single image dehazing method. In this work they reduced the image dehazing problem to an \textbf{paired} image-to-image translation problem and proposed an enhanced Pix2pix Dehazing network based on a generative adversarial network. This network contains generators, discriminators, and two enhancing blocks to produce a realistic dehazed image on the fine scale. 

The enhancer contains two enhancing blocks based on the receptive field model, which reinforces the dehazing effect in both color and details. The GAN is jointly trained with the enhancer.

\subsection{CycleDehaze}
CycleDehaze~\cite{zhu2017unpaired} is an end-to-end single image dehazing method which does not require pairs of hazy and corresponding ground truth images for training,~\emph{i.e.} they train the network by feeding clean and hazy images in an unpaired manner. 

This method enhances CycleGAN formulation by combining cycle-consistency and perceptual loss to improve the quality of textural information recovery and generate more visually pleasing and realistic haze-free images. 

%This method is also a learning-based method and does not rely on estimation of the atmospheric scattering model parameters.

\subsection{Dehaze-GLCGAN}
Dehaze-GLCGAN~\cite{anvari2020dehaze} is another GAN-based single image dehazing method based on a global-local dehazing mechanism. They cast the image dehazing problem to an \textbf{unpaired} image-to-image translation problem which means the pairs of hazy images and the corresponding ground truth images are not required for training process similar to CycleDehaze~\cite{engin2018cycle}. 

They proposed a dehazing Global-Local Cycle-consistent Generative Adversarial Network. Generator network of Dehaze-GLCGAN combines an encoder-decoder architecture with residual blocks to better recover the haze-free scene. They also proposed a global-local discriminator structure to deal with spatially varying haze.

\section{Results and Discussion}
In this section, we present the quantitative and qualitative evaluation results and then discuss the performance of the dehazing methods over Sun-Haze dataset.

\subsection{Quantitative Evaluation}
In this section, the Sun-Haze dataset is used to perform a comprehensive quantitative evaluation of several state-of-the-art single image dehazing methods, as described in Section~\ref{existing}.

We used several metrics to evaluate these dehazing methods, including reference-based and no-reference-based metrics. These metrics are:

%\begin{table}[t]
%\ra{1.2}
%\caption{Results over our Sun-Haze. For this analysis, the original image is used as the ground truth/haze free.}
%\label{no_experts}
%\centering
%\setlength{\tabcolsep}{6pt}
%\begin{tabular}{llllll}
%\toprule
% Method & $\uparrow$PSNR & $\uparrow$SSIM & $\downarrow$NIQE & $\downarrow$CIEDE2000 & $\downarrow$PI\\\toprule 
%\rowcolor{lightgray} Hazy image & 17.25 & 0.879 & 3.59 & 16.76  \\
%DCP~\cite{he2010single} & 11.40 & 0.686 & 5.35 & 36.10& 3.71 \\
%MSCNN~\cite{ren2016single} & \textbf{19.10} & \textbf{0.867} & 4.06 & %\textbf{18.44} &3.25 \\
%DehazeNet~\cite{cai2016dehazenet} & 17.78 & 0.814 & 4.09 &23.95& 3.25    \\
%AODNet~\cite{li2017aod} & 16.89 & 0.782 & \textbf{3.93} &26.63 & 3.02     \\
%EPDN~\cite{qu2019enhanced} & 17.96 & 0.857 & 4.13  & 22.72 & 3.041 \\
%CycleGAN~\cite{zhu2017unpaired} & 14.0357  & 0.7599 & 3.5779   \\
%GLCGAN~\cite{anvari2020dehaze} & 14.57 & 0.819 & 4.09 & 24.46 &\textbf{2.93} \\
%CycleDehaze~\cite{engin2018cycle} & 16.39 & 0.834 & 4.60 & 23.46&4.08\\
%\hline
%\end{tabular}
%\end{table}

\begin{table}[!t]
\ra{1.3}
\centering
\caption{Results over Sun-Haze dataset. We performed separate analysis for different ground truth images. The images retouched by 5 experts and the original image before retouch are considered as ground truth/haze free. We also present results for the no-reference metrics.}
    \resizebox{1.0\textwidth}{!}{\begin{tabular}{clcccccccc}\toprule
    Ground truth & Metric & DCP & MSCNN & Dehazenet & AOD-Net & EPDN & Dehaze-GLCGAN & CycleDehaze\\
    \toprule
     & PSNR & 11.01 & \textbf{16.48} & 15.62 & 14.83 & 15.88 & 14.38 & 15.53\\
    \textbf{Expert A} & SSIM  & 0.641 & 0.773 & 0.733 & 0.698 & 0.784 & \textbf{0.789} & 0.778\\
     & CIEDE & 34.76 & \textbf{23.46} & 27.55 & 30.26 & 26.29 & 24.37 & 26.62\\\hline
    & PSNR & 11.28 & 16.33 & 15.13 & 14.15 & 14.96 & 15.12 & 15.38\\
    \textbf{Expert B} & SSIM & 0.655 & 0.763 & 0.709 & 0.676 & 0.761 & \textbf{0.801} & 0.763\\
     & CIEDE & 32.66 & 25.43 & 30.90 & 34.33 & 31.64 & \textbf{22.03} & 26.49\\\hline
    & PSNR & 11.32 & \textbf{16.57} & 15.49 & 14.49 & 15.44 & 14.74 & 15.23 \\
    \textbf{Expert C} & SSIM & 0.643 & 0.746 & 0.703 & 0.670 & 0.756 & \textbf{0.782} & 0.737\\
     & CIEDE & 33.22 & 24.57 & 28.84 & 31.93 & 28.73 & \textbf{23.98} &28.98 \\\hline
    & PSNR & 11.43 & 14.91 & 13.75 & 12.82 & 13.55 & \textbf{14.93} & 14.35\\
    \textbf{Expert D} & SSIM & 0.649 & 0.722 & 0.667 & 0.632 & 0.713 & \textbf{0.781} & 0.728\\
     & CIEDE & 30.87 & 29.10 & 34.76 & 38.87 & 36.28 & \textbf{22.36} & 28.67\\\hline
    & PSNR &11.32 & 15.27 & 13.86 & 12.99 & 13.56 & \textbf{15.32} & 14.84\\
    \textbf{Expert E} & SSIM & 0.640 & 0.719 & 0.660 & 0.626 & 0.704 & \textbf{0.780} & 0.733\\
     & CIEDE &33.07 & 28.56 & 34.38 & 38.11 & 35.88 & \textbf{22.30} & 28.13\\\hline
    & PSNR &11.40 & \textbf{19.10} & 17.78 & 16.89 & 17.96 & 14.57 & 16.39 \\
    \textbf{Original image} & SSIM & 0.686 & \textbf{0.867} & 0.814 & 0.782  & 0.857 & 0.810 & 0.834 \\
     & CIEDE & 36.10 & \textbf{18.44} & 23.95 & 26.63 & 22.72 & 24.46 & 23.46\\\toprule
    & PSNR & 11.34 & \colorbox{YellowGreen}{\textbf{17.51}} & \colorbox{pink}{16.28} & 15.37 & \colorbox{flavescent}{16.32} & 14.73 & 15.73 \\
    \textbf{Avgerage} & SSIM & 0.651 & \colorbox{flavescent}{0.765} & 0.721 & 0.680  & \colorbox{pink}{0.763} & \colorbox{YellowGreen}{\textbf{0.799}} & 0.762 \\
     & CIEDE & 33.45 & \colorbox{flavescent}{24.93} & 30.06 & 33.36 & 30.26 &  \colorbox{YellowGreen}{\textbf{23.25}} & \colorbox{pink}{27.06}\\\toprule\toprule
    \textbf{No reference} & NIQE & 5.35 & \colorbox{flavescent}{4.06} & \colorbox{pink}{4.08} & \colorbox{YellowGreen}{\textbf{3.93}} & 4.13 & 4.09 & 4.60 \\
    \textbf{No reference} & PI & 3.71 & 3.25 & 3.25 & \colorbox{flavescent}{3.02}  & \colorbox{pink}{3.04} & \colorbox{YellowGreen}{\textbf{2.93}} & 4.08 \\\toprule
\end{tabular}}
    \label{experts}
\end{table}

\begin{itemize}
    \item \textbf{PSNR}: This metric provides a pixel-wise evaluation and is capable of indicating the effectiveness of haze removal. It measures the ratio between the maximum possible value of a signal and the power of distorting noise that affects the quality of its representation. 
    \item \textbf{SSIM}: It is a perceptual metric for measuring the structural similarity between two images.
    \item \textbf{CIEDE2000}: It measures the color difference between hazy and dehazed images.
    \item \textbf{NIQE}: It is a well-known no-reference image quality assessment metric for evaluating real image restoration without requiring the ground-truth.
    \item \textbf{PI}: Perceptual Index measures the quality of the reconstructed images based on human perception. A lower perceptual index indicates better perceptual quality. 
    The perceptual quality of an image is the degree to which the image looks like a natural image. This metric is a no-reference quality measurement metric which means that it does not require a ground truth image.
\end{itemize}

%The  perceptual  quality  of  an  image x is  the  degree  to which it looks like a natural image, and has nothing to do with  its  similarity  to  any  reference  image.

The larger values of PSNR, SSIM and the smaller values of CIEDE2000, NIQE, and PI indicate better dehazing and perceptual quality.
 
Table~\ref{experts} shows the quantitative results of evaluating the dehazing methods in terms of PSNR and SSIM, CIEDE2000, NIQE, and PI. We conducted multiple experiments to evaluate the dehazing and generalization capability of the dehazing techniques. 

%In the first experiment we used our Sun-Haze dataset as the hazy images and the original images (before retouch) as the ground truth(the Original image row in Table~\ref{experts}).

We used our Sun-Haze dataset as the hazy images and the retouched images by five experts as well as the original image as the ground truth for each experiment. The best results are depicted in bold. We also highlighted the top three best results. The green highlights represent the best results, yellow the second best and pink the third best results.

As one can observe, on average MSCNN achieved best PSNR of 17.51 and the second best SSIM and CIEDE2000. GLCGAN outperformed other methods in terms of SSIM, CIEDE2000, and PI by a very small margin. AOD-Net achieved the best NIQE of 3.93 which is very comparable with the second and third best results.

As you can see, no method is superior to other methods in all five measurements and the best results are only slightly better than the second and third best results. This suggests that current methods can not generalize well to remove haze caused by sunlight.

%Table~\ref{experts} shows the quantitative results over our Sun-Haze, but for this analysis we used the images retouched by five experts as the ground truth images. First column shows the metrics for the hazy images themselves that serves as a baseline for this analysis. The best results are depicted in bold.

%On average, MSCNN method outperformed other methods in terms of PSNR, and GLCGAN achieved the highest SSIM and the lowest CIEDE2000.

%the corresponding ground truth (rightmost column)
\begin{figure}
    \centering
    \begin{subfigure}[b]{0.166\textwidth}
        \centering
        {Hazy image}
        \includegraphics[width=0.95\textwidth]{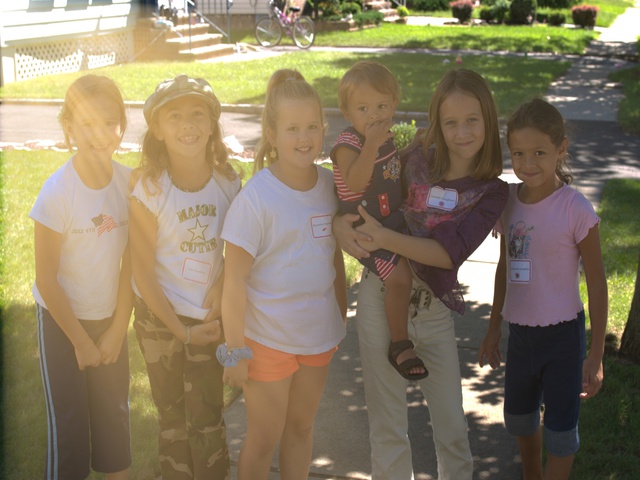}
        \label{fig:hazy}
    \end{subfigure}
        \hspace*{-0.05in}%
        \begin{subfigure}[b]{0.166\textwidth}
        \centering
        {AOD-Net~\cite{li2017aod}}
        \includegraphics[width=0.95\textwidth]{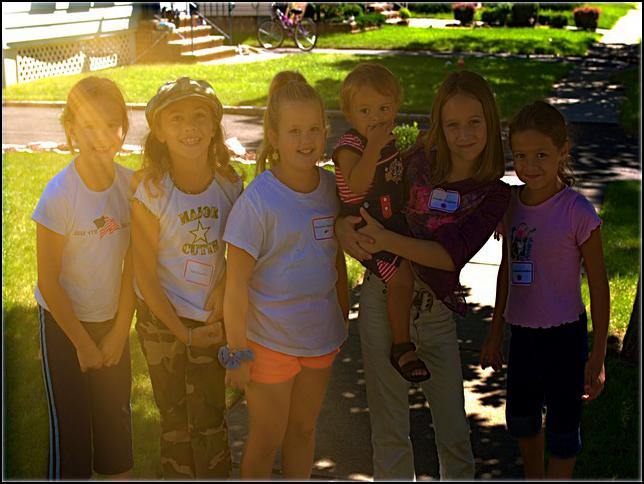}
    \end{subfigure}
        \hspace*{-0.05in}%
    \begin{subfigure}[b]{0.166\textwidth}
        \centering
        {GLCGAN~\cite{anvari2020dehaze}}
        \includegraphics[width=0.95\textwidth]{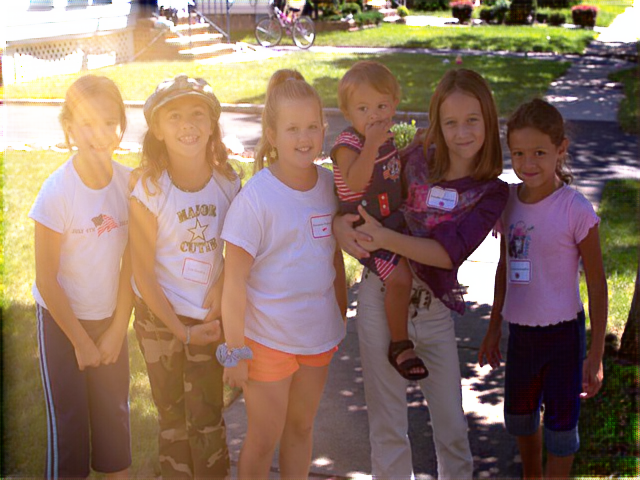}
        \label{fig:hazy}
    \end{subfigure}
    \hspace*{-0.05in}%
    \begin{subfigure}[b]{0.166\textwidth}  
        \centering 
        {DehazeNet~\cite{cai2016dehazenet}}
        \includegraphics[width=0.95\textwidth]{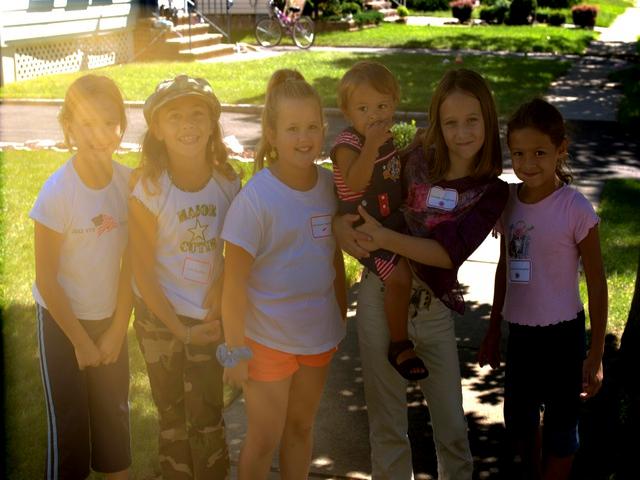}
        \label{fig:aodnet}
    \end{subfigure}
    \hspace*{-0.05in}%
    \begin{subfigure}[b]{0.166\textwidth}   
        \centering 
        {MSCNN~\cite{ren2016single}}
        \includegraphics[width=0.95\textwidth]{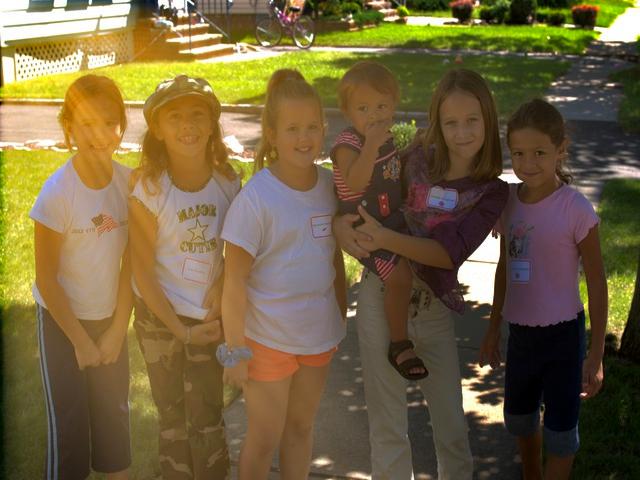}
        \label{fig:epdn}
    \end{subfigure}
    \hspace*{-0.05in}%
    \begin{subfigure}[b]{0.166\textwidth}   
        \centering 
        {EPDN~\cite{qu2019enhanced}}
        \includegraphics[width=0.95\textwidth]{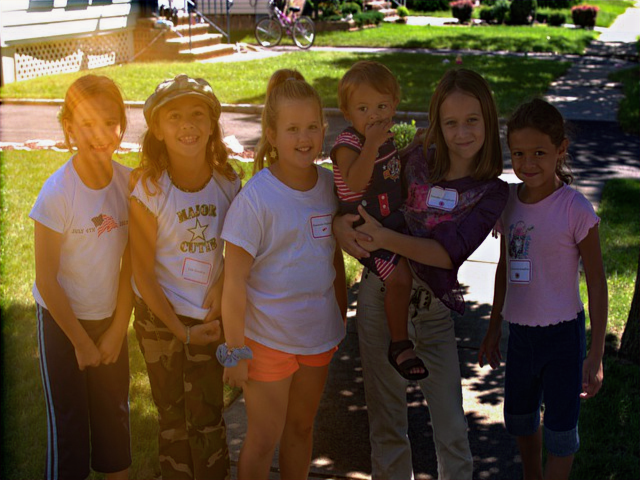}
        \label{fig:ours1}
    \end{subfigure}
    \hspace*{-0.05in}%
    \begin{subfigure}[b]{0.166\textwidth}   
        \centering 
        {Original}
        \includegraphics[width=0.95\textwidth]{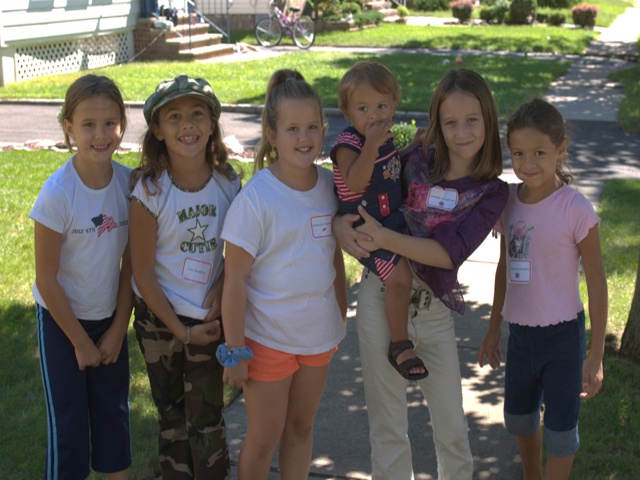}
        \label{fig:ours1}
    \end{subfigure}
    \hspace*{-0.05in}%
    \begin{subfigure}[b]{0.166\textwidth}   
        \centering 
        {Expert A}
        \includegraphics[width=0.95\textwidth]{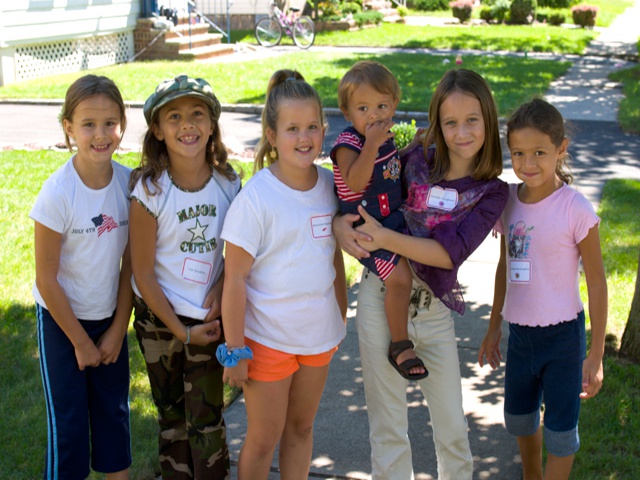}
        \label{fig:ours1}
    \end{subfigure}
        \hspace*{-0.05in}%
    \begin{subfigure}[b]{0.166\textwidth}   
        \centering 
        {Expert B}
        \includegraphics[width=0.95\textwidth]{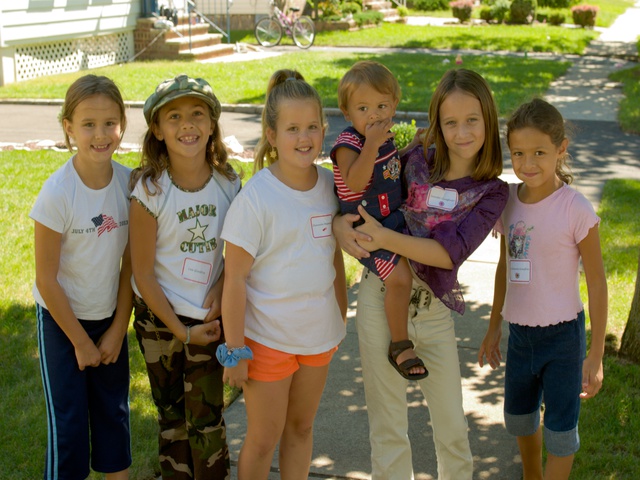}
        \label{fig:ours1}
    \end{subfigure}
    \hspace*{-0.05in}%
    \begin{subfigure}[b]{0.166\textwidth}   
        \centering 
        {Expert C}
        \includegraphics[width=0.95\textwidth]{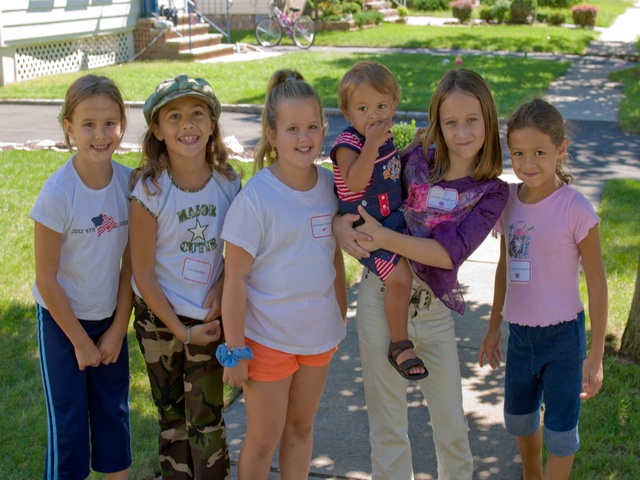}
        \label{fig:ours1}
    \end{subfigure}
        \hspace*{-0.05in}%
    \begin{subfigure}[b]{0.166\textwidth}   
        \centering 
        {Expert D}
        \includegraphics[width=0.95\textwidth]{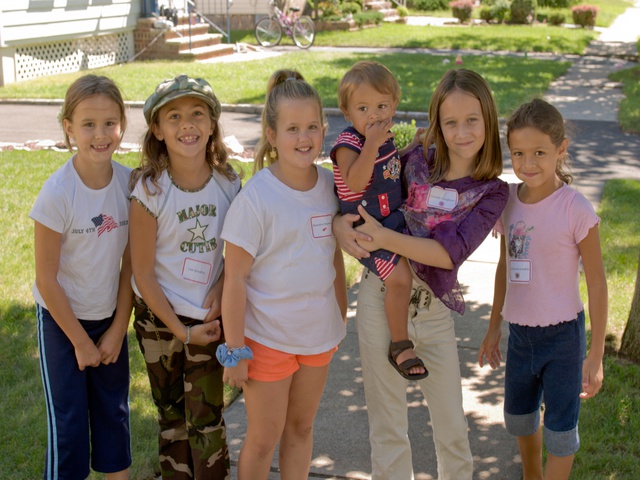}
        \label{fig:ours1}
    \end{subfigure}
    \hspace*{-0.05in}%
    \begin{subfigure}[b]{0.166\textwidth}   
        \centering 
        {Expert E}
        \includegraphics[width=0.95\textwidth]{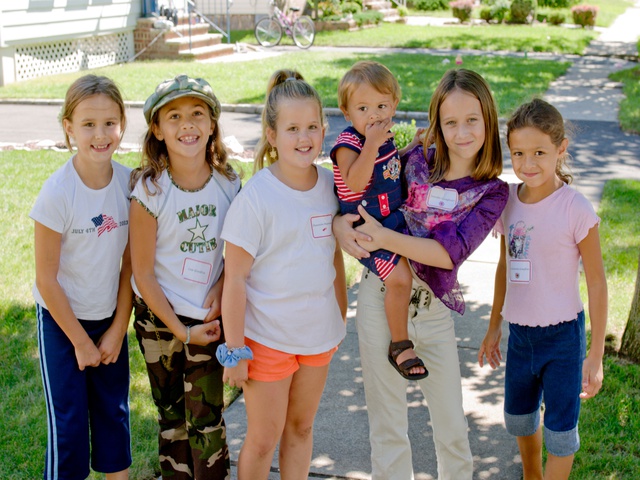}
        \label{fig:ours1}
    \end{subfigure}
    
    %%%%%%%%%%%%%%%%%%%%%%%%%%%%%%
    %\hspace*{0.05in}%
    \vspace*{0.1in}
    \begin{subfigure}[b]{0.166\textwidth}
        \centering
        {Hazy image}
        \includegraphics[width=0.95\textwidth]{figs/dataset/a0814-MB_070908_062_2.jpg}
        \label{fig:hazy}
    \end{subfigure}
        \hspace*{-0.05in}%
        \begin{subfigure}[b]{0.166\textwidth}
        \centering
        {AOD-Net}
        \includegraphics[width=0.95\textwidth]{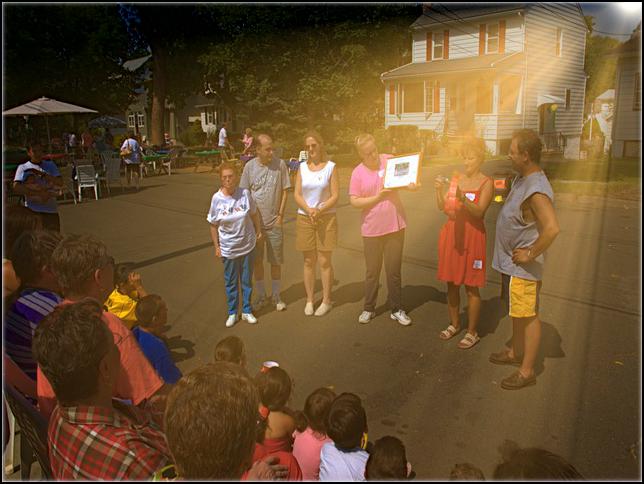}
    \end{subfigure}
        \hspace*{-0.05in}%
    \begin{subfigure}[b]{0.166\textwidth}
        \centering
        {GLCGAN}
        \includegraphics[width=0.95\textwidth]{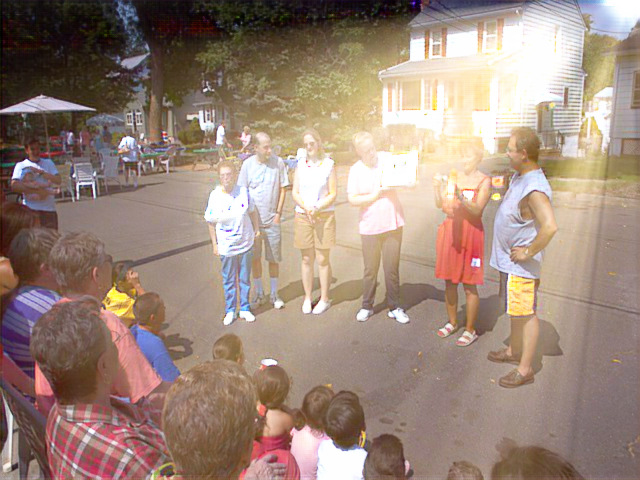}
        \label{fig:hazy}
    \end{subfigure}
    \hspace*{-0.05in}%
    \begin{subfigure}[b]{0.166\textwidth}  
        \centering 
        {DehazeNet}
        \includegraphics[width=0.95\textwidth]{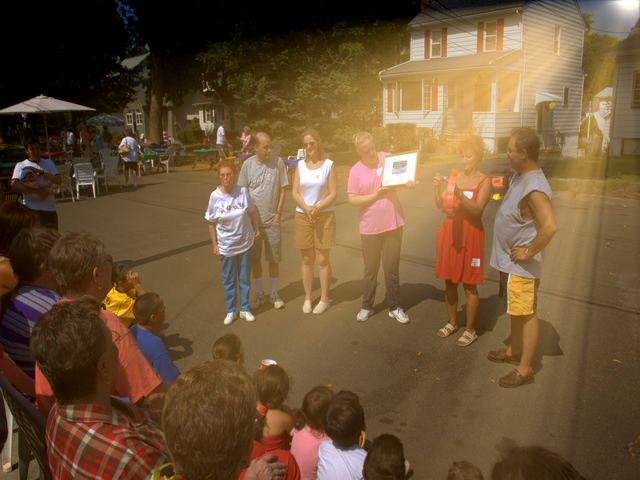}
        \label{fig:aodnet}
    \end{subfigure}
    \hspace*{-0.05in}%
    \begin{subfigure}[b]{0.166\textwidth}   
        \centering 
        {MSCNN}
        \includegraphics[width=0.95\textwidth]{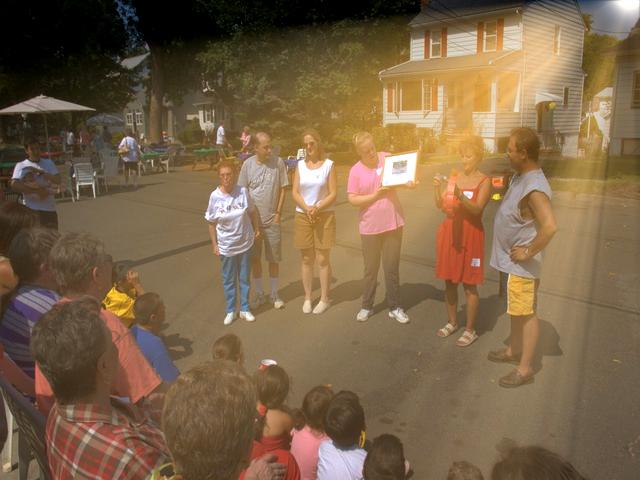}
        \label{fig:epdn}
    \end{subfigure}
    \hspace*{-0.05in}%
    \begin{subfigure}[b]{0.166\textwidth}   
        \centering 
        {EPDN}
        \includegraphics[width=0.95\textwidth]{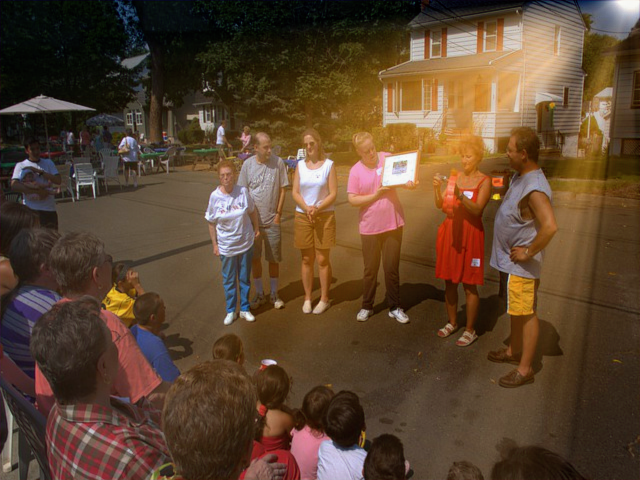}
        \label{fig:ours1}
    \end{subfigure}
    \hspace*{-0.05in}%
    \begin{subfigure}[b]{0.166\textwidth}   
        \centering 
        {Original}
        \includegraphics[width=0.95\textwidth]{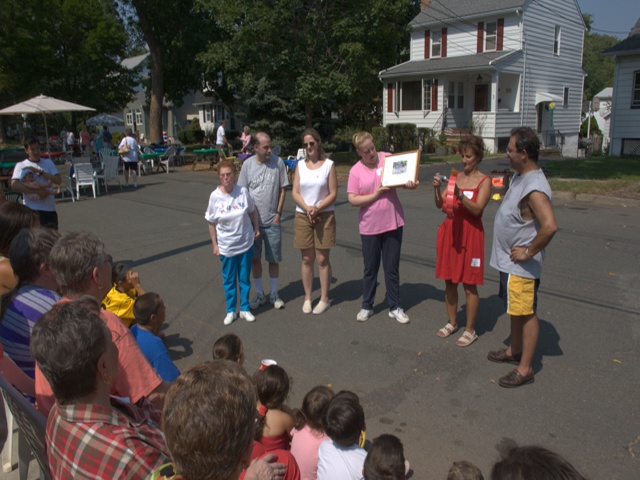}
        \label{fig:ours1}
    \end{subfigure}
    \hspace*{-0.05in}%
    \begin{subfigure}[b]{0.166\textwidth}   
        \centering 
        {Expert A}
        \includegraphics[width=0.95\textwidth]{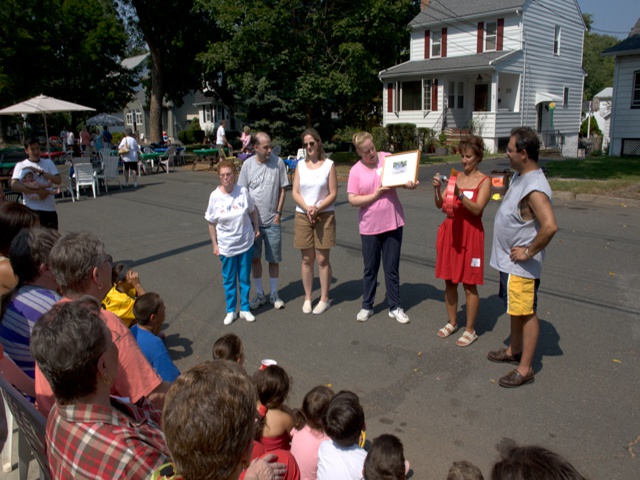}
        \label{fig:ours1}
    \end{subfigure}
        \hspace*{-0.05in}%
    \begin{subfigure}[b]{0.166\textwidth}   
        \centering 
        {Expert B}
        \includegraphics[width=0.95\textwidth]{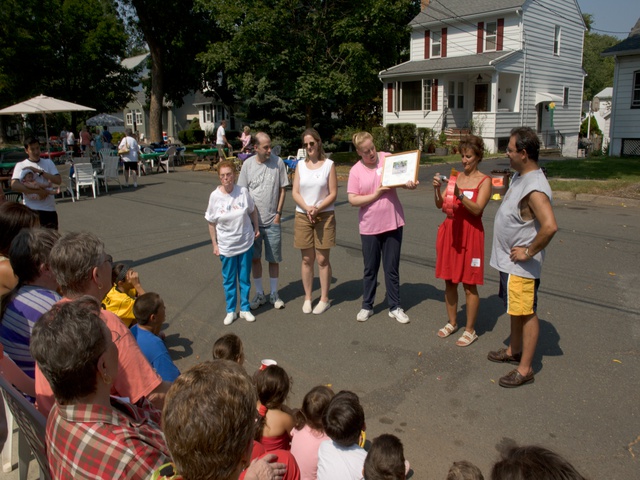}
        \label{fig:ours1}
    \end{subfigure}
    \hspace*{-0.05in}%
    \begin{subfigure}[b]{0.166\textwidth}   
        \centering 
        {Expert C}
        \includegraphics[width=0.95\textwidth]{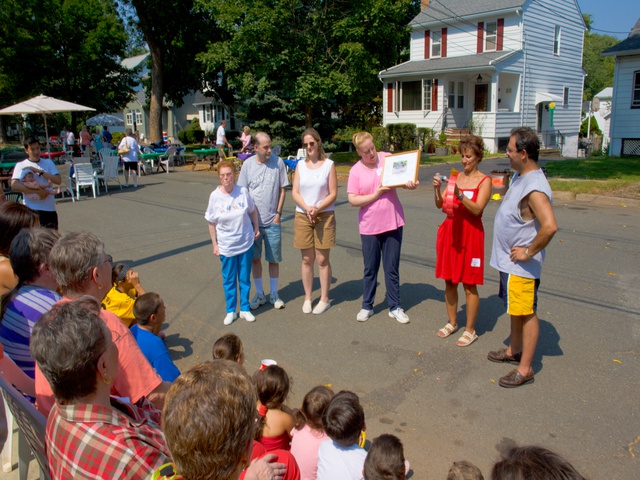}
        \label{fig:ours1}
    \end{subfigure}
        \hspace*{-0.05in}%
    \begin{subfigure}[b]{0.166\textwidth}   
        \centering 
        {Expert D}
        \includegraphics[width=0.95\textwidth]{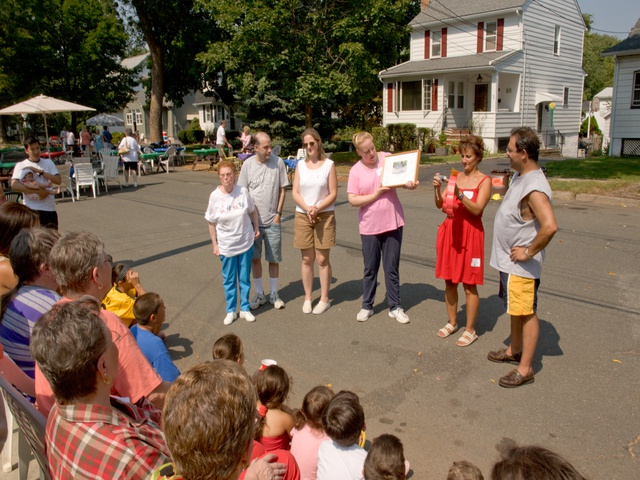}
        \label{fig:ours1}
    \end{subfigure}
    \hspace*{-0.05in}%
    \begin{subfigure}[b]{0.166\textwidth}   
        \centering 
        {Expert E}
        \includegraphics[width=0.95\textwidth]{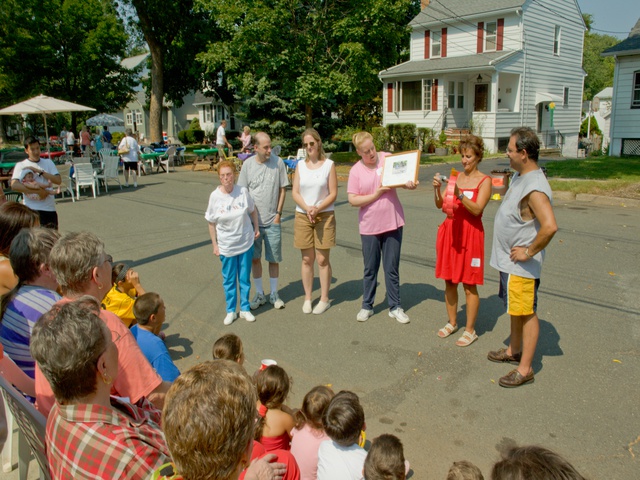}
        \label{fig:ours1}
    \end{subfigure}

    %%%%%%%%%%%%%%%%%%%%%%%%%%%%%%
    \vspace*{0.1in}
    \begin{subfigure}[b]{0.166\textwidth}
        \centering
        {Hazy image}
        \includegraphics[width=0.95\textwidth]{figs/dataset/a0900-jmac_MG_7376.jpg}
        \label{fig:hazy}
    \end{subfigure}
        \hspace*{-0.05in}%
        \begin{subfigure}[b]{0.166\textwidth}
        \centering
        {AOD-Net}
        \includegraphics[width=0.95\textwidth]{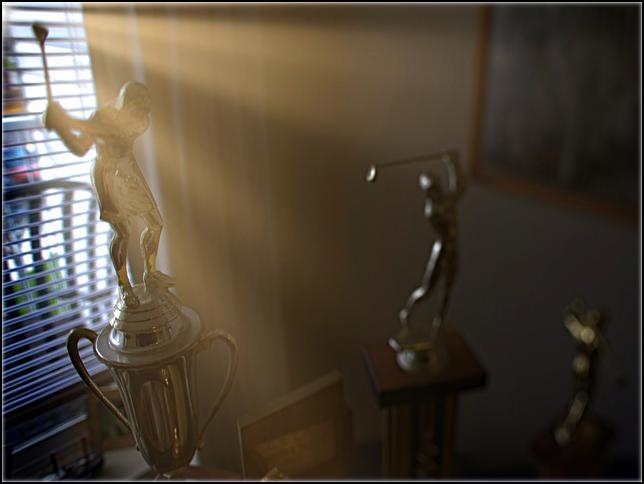}
    \end{subfigure}
        \hspace*{-0.05in}%
    \begin{subfigure}[b]{0.166\textwidth}
        \centering
        {GLCGAN}
        \includegraphics[width=0.95\textwidth]{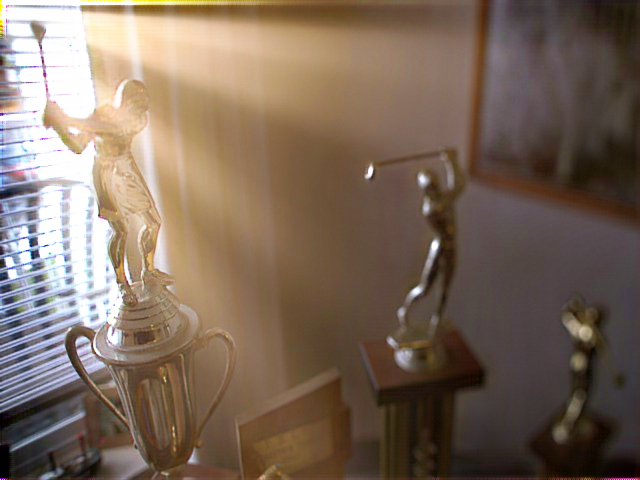}
        \label{fig:hazy}
    \end{subfigure}
    \hspace*{-0.05in}%
    \begin{subfigure}[b]{0.166\textwidth}  
        \centering 
        {DehazeNet}
        \includegraphics[width=0.95\textwidth]{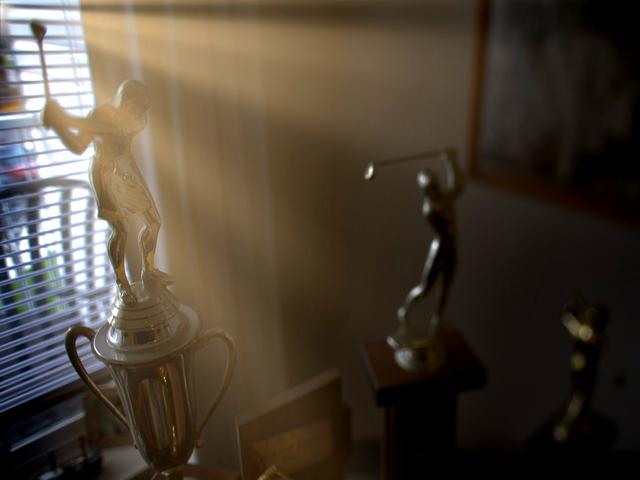}
        \label{fig:aodnet}
    \end{subfigure}
    \hspace*{-0.05in}%
    \begin{subfigure}[b]{0.166\textwidth}   
        \centering 
        {MSCNN}
        \includegraphics[width=0.95\textwidth]{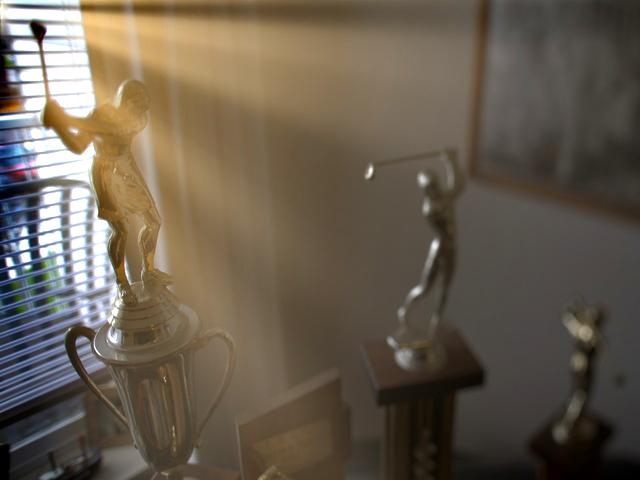}
        \label{fig:epdn}
    \end{subfigure}
    \hspace*{-0.05in}%
    \begin{subfigure}[b]{0.166\textwidth}   
        \centering 
        {EPDN}
        \includegraphics[width=0.95\textwidth]{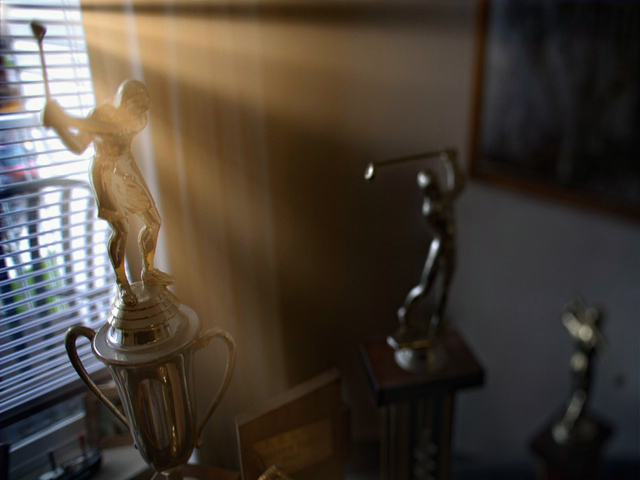}
        \label{fig:ours1}
    \end{subfigure}
    \hspace*{-0.05in}%
    \begin{subfigure}[b]{0.166\textwidth}   
        \centering 
        {Original}
        \includegraphics[width=0.95\textwidth]{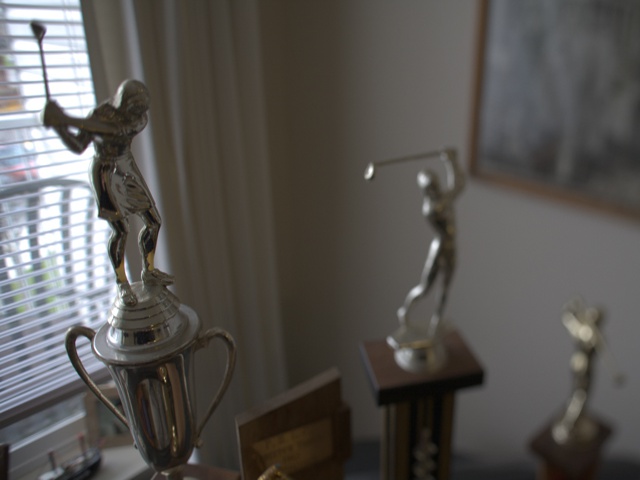}
        \label{fig:ours1}
    \end{subfigure}
    \hspace*{-0.05in}%
    \begin{subfigure}[b]{0.166\textwidth}   
        \centering 
        {Expert A}
        \includegraphics[width=0.95\textwidth]{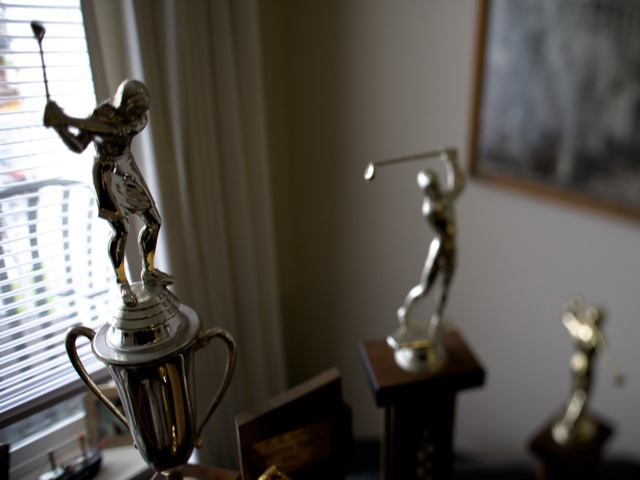}
        \label{fig:ours1}
    \end{subfigure}
        \hspace*{-0.05in}%
    \begin{subfigure}[b]{0.166\textwidth}   
        \centering 
        {Expert B}
        \includegraphics[width=0.95\textwidth]{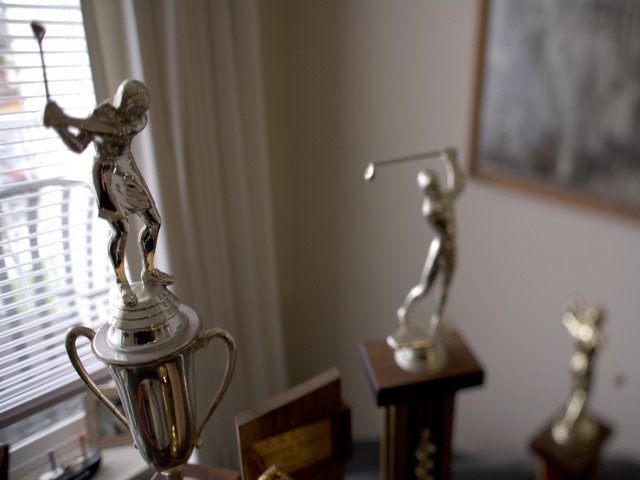}
        \label{fig:ours1}
    \end{subfigure}
    \hspace*{-0.05in}%
    \begin{subfigure}[b]{0.166\textwidth}   
        \centering 
        {Expert C}
        \includegraphics[width=0.95\textwidth]{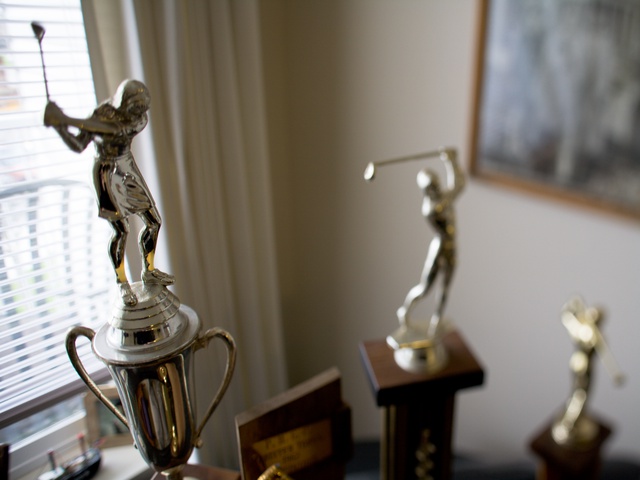}
        \label{fig:ours1}
    \end{subfigure}
        \hspace*{-0.05in}%
    \begin{subfigure}[b]{0.166\textwidth}   
        \centering 
        {Expert D}
        \includegraphics[width=0.95\textwidth]{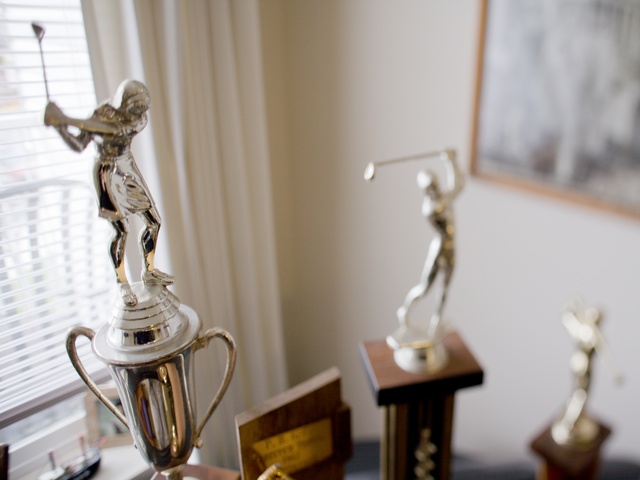}
        \label{fig:ours1}
    \end{subfigure}
    \hspace*{-0.05in}%
    \begin{subfigure}[b]{0.166\textwidth}   
        \centering 
        {Expert E}
        \includegraphics[width=0.95\textwidth]{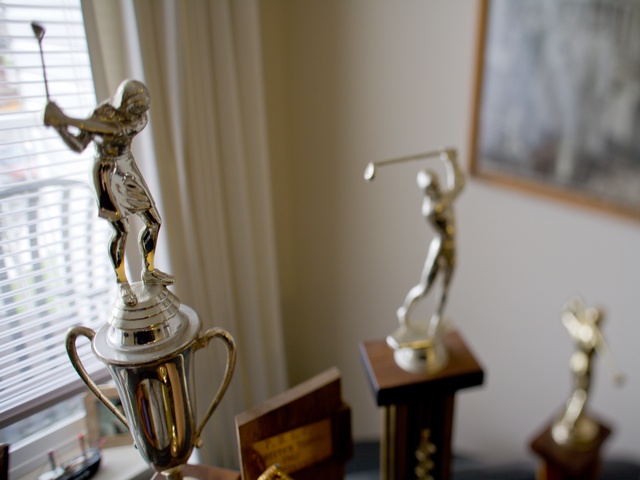}
        \label{fig:ours1}
    \end{subfigure}
    
    %%%%%%%%%%%%%%%%%%%%%%%%%%%%%%
    \vspace*{0.1in}
    \begin{subfigure}[b]{0.166\textwidth}
        \centering
        {Hazy image}
        \includegraphics[width=0.95\textwidth]{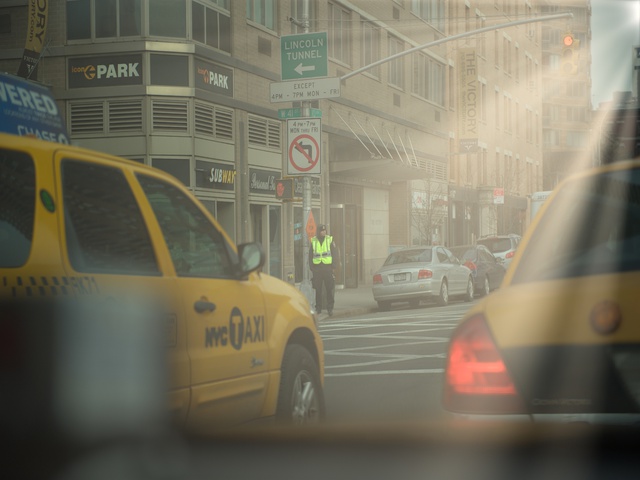}
        \label{fig:hazy}
    \end{subfigure}
        \hspace*{-0.05in}%
        \begin{subfigure}[b]{0.166\textwidth}
        \centering
        {AOD-Net}
        \includegraphics[width=0.95\textwidth]{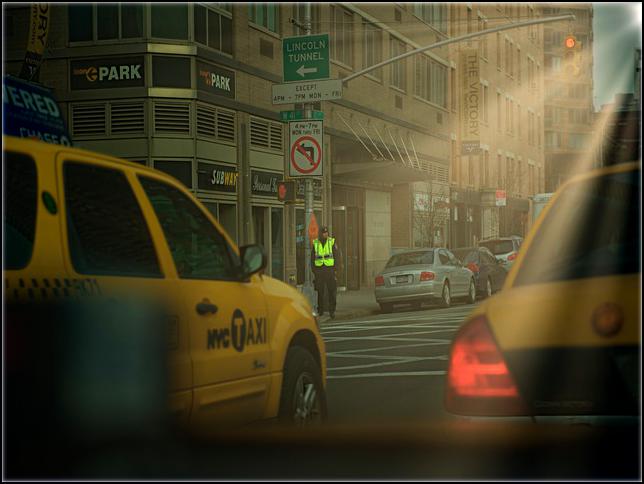}
    \end{subfigure}
        \hspace*{-0.05in}%
    \begin{subfigure}[b]{0.166\textwidth}
        \centering
        {GLCGAN}
        \includegraphics[width=0.95\textwidth]{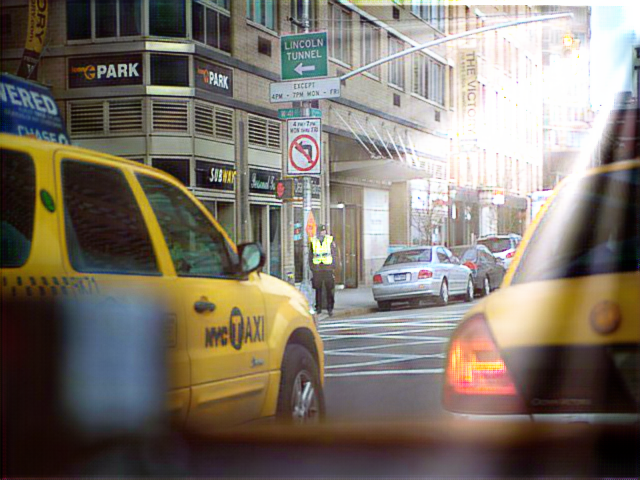}
        \label{fig:hazy}
    \end{subfigure}
    \hspace*{-0.05in}%
    \begin{subfigure}[b]{0.166\textwidth}  
        \centering 
        {DehazeNet}
        \includegraphics[width=0.95\textwidth]{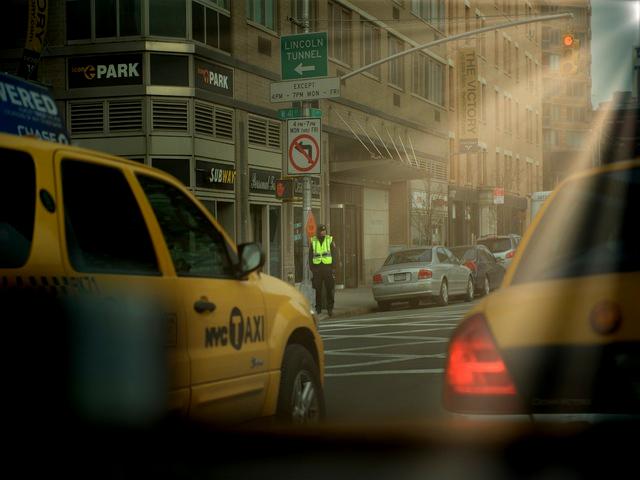}
        \label{fig:aodnet}
    \end{subfigure}
    \hspace*{-0.05in}%
    \begin{subfigure}[b]{0.166\textwidth}   
        \centering 
        {MSCNN}
        \includegraphics[width=0.95\textwidth]{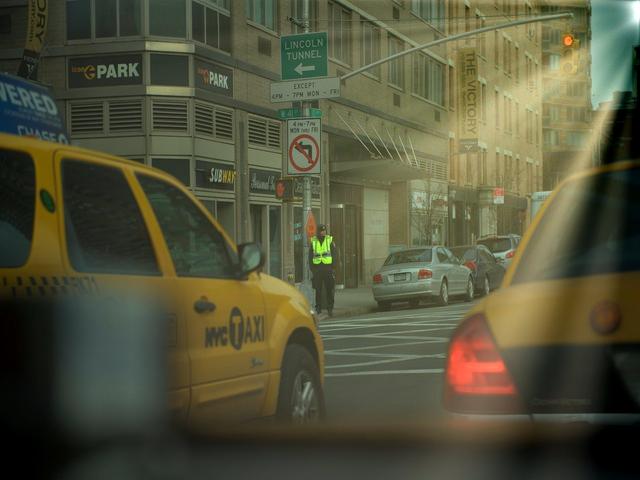}
        \label{fig:epdn}
    \end{subfigure}
    \hspace*{-0.05in}%
    \begin{subfigure}[b]{0.166\textwidth}   
        \centering 
        {EPDN}
        \includegraphics[width=0.95\textwidth]{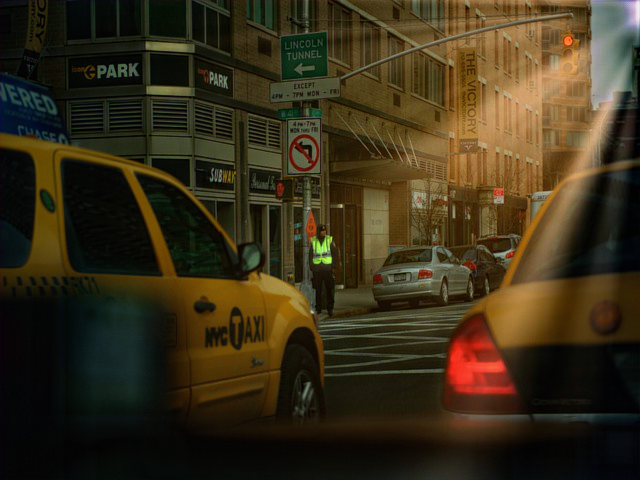}
        \label{fig:ours1}
    \end{subfigure}
    \hspace*{-0.05in}%
    \begin{subfigure}[b]{0.166\textwidth}   
        \centering 
        {Original}
        \includegraphics[width=0.95\textwidth]{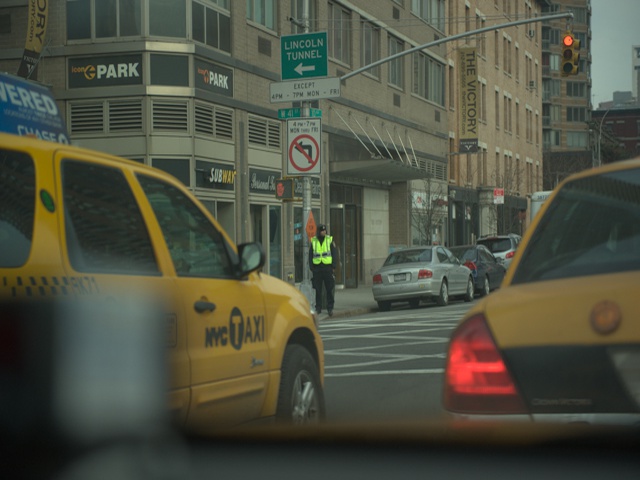}
        \label{fig:ours1}
    \end{subfigure}
    \hspace*{-0.05in}%
    \begin{subfigure}[b]{0.166\textwidth}   
        \centering 
        {Expert A}
        \includegraphics[width=0.95\textwidth]{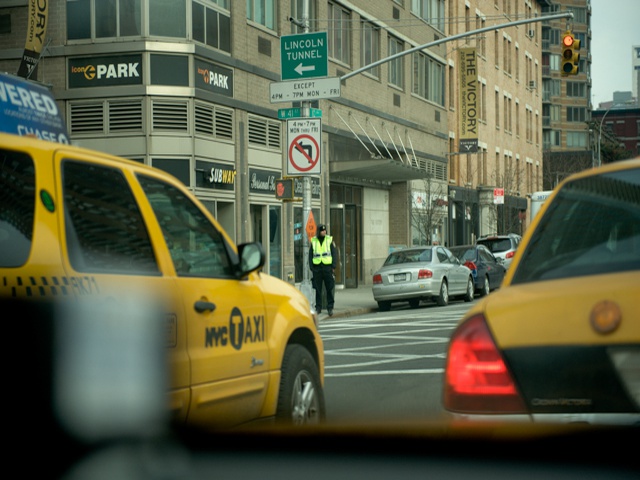}
        \label{fig:ours1}
    \end{subfigure}
        \hspace*{-0.05in}%
    \begin{subfigure}[b]{0.166\textwidth}   
        \centering 
        {Expert B}
        \includegraphics[width=0.95\textwidth]{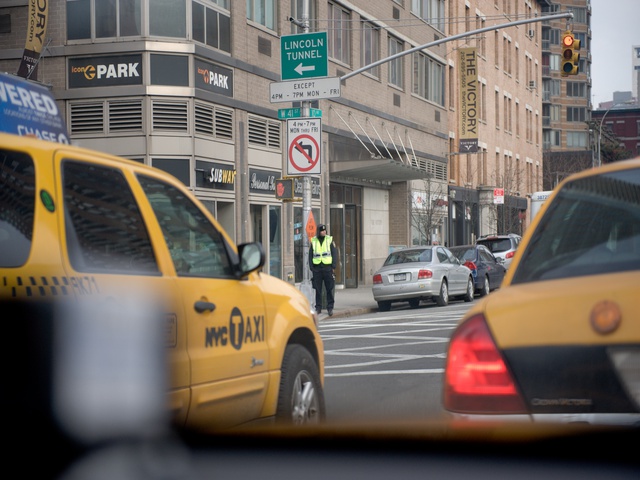}
        \label{fig:ours1}
    \end{subfigure}
    \hspace*{-0.05in}%
    \begin{subfigure}[b]{0.166\textwidth}   
        \centering 
        {Expert C}
        \includegraphics[width=0.95\textwidth]{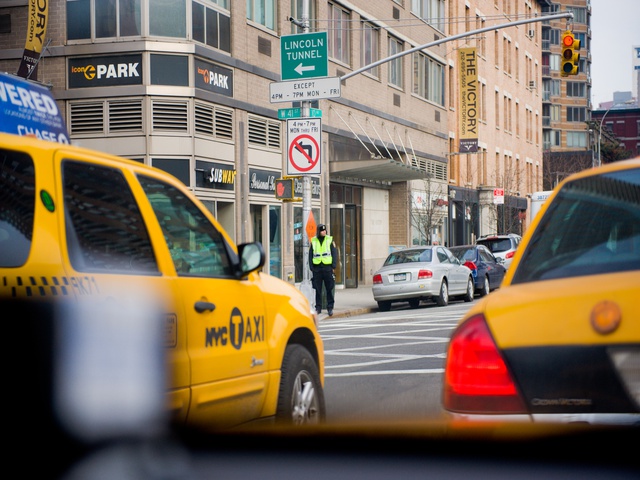}
        \label{fig:ours1}
    \end{subfigure}
        \hspace*{-0.05in}%
    \begin{subfigure}[b]{0.166\textwidth}   
        \centering 
        {Expert D}
        \includegraphics[width=0.95\textwidth]{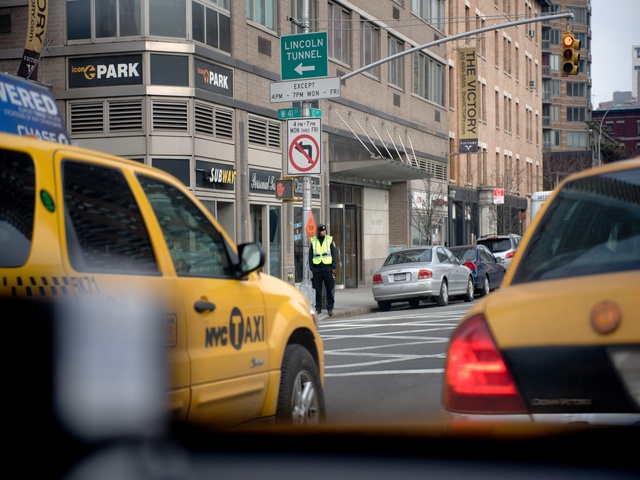}
        \label{fig:ours1}
    \end{subfigure}
    \hspace*{-0.05in}%
    \begin{subfigure}[b]{0.166\textwidth}   
        \centering 
        {Expert E}
        \includegraphics[width=0.95\textwidth]{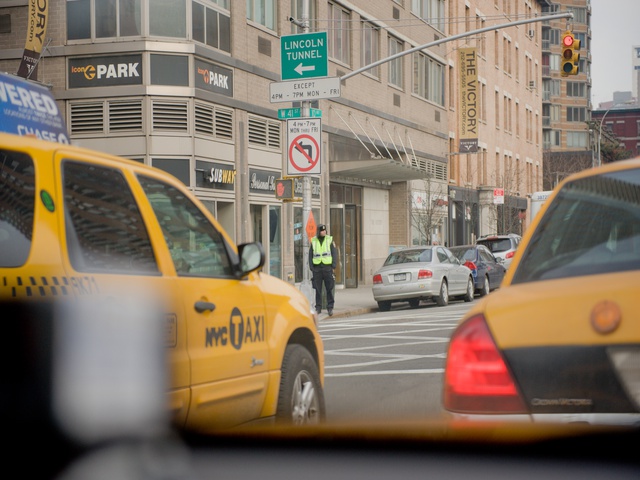}
        \label{fig:ours1}
    \end{subfigure}
    
        %%%%%%%%%%%%%%%%%%%%%%%%%%%%%%
    \vspace*{0.1in}
    \begin{subfigure}[b]{0.166\textwidth}
        \centering
        {Hazy image}
        \includegraphics[width=0.95\textwidth]{figs/sunset/input/a0773.png}
        \label{fig:hazy}
    \end{subfigure}
        \hspace*{-0.05in}%
        \begin{subfigure}[b]{0.166\textwidth}
        \centering
        {AOD-Net}
        \includegraphics[width=0.95\textwidth]{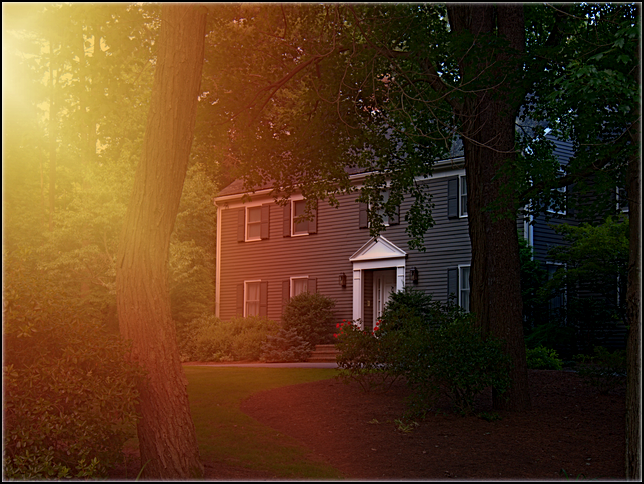}
    \end{subfigure}
        \hspace*{-0.05in}%
    \begin{subfigure}[b]{0.166\textwidth}
        \centering
        {GLCGAN}
        \includegraphics[width=0.95\textwidth]{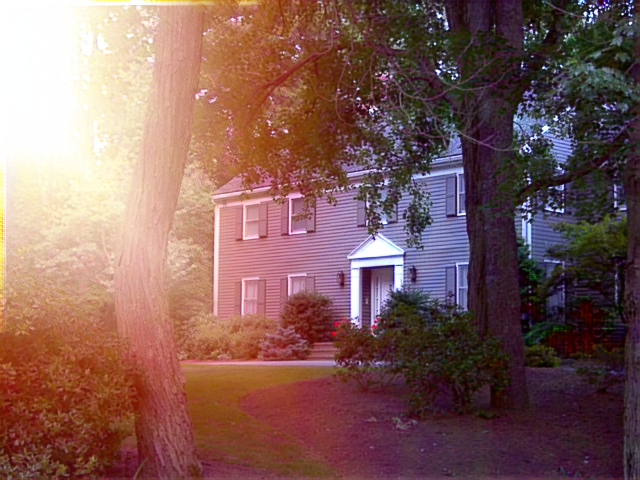}
        \label{fig:hazy}
    \end{subfigure}
    \hspace*{-0.05in}%
    \begin{subfigure}[b]{0.166\textwidth}  
        \centering 
        {DehazeNet}
        \includegraphics[width=0.95\textwidth]{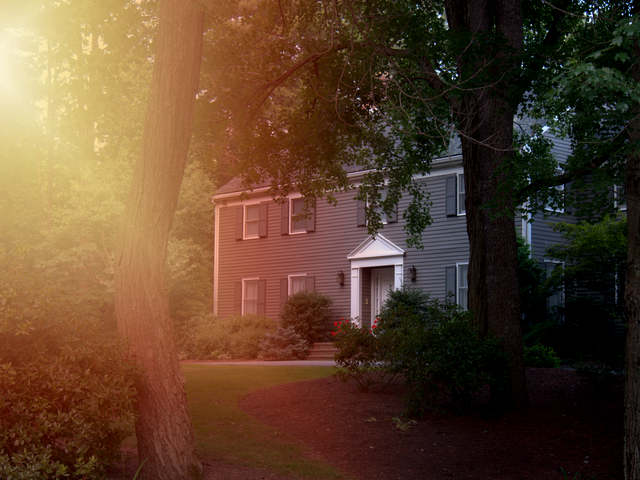}
        \label{fig:aodnet}
    \end{subfigure}
    \hspace*{-0.05in}%
    \begin{subfigure}[b]{0.166\textwidth}   
        \centering 
        {MSCNN}
        \includegraphics[width=0.95\textwidth]{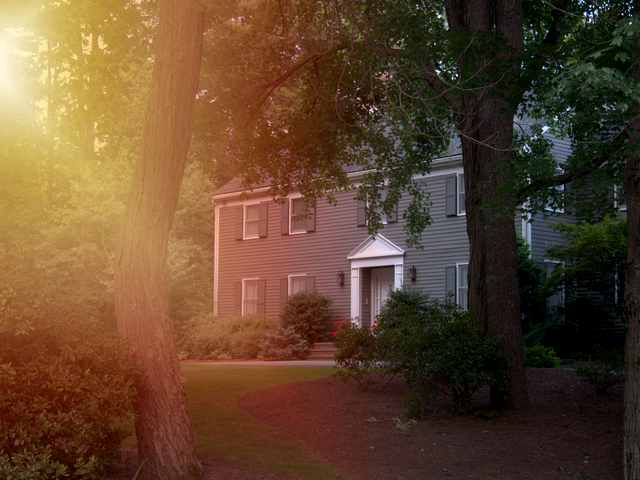}
        \label{fig:epdn}
    \end{subfigure}
    \hspace*{-0.05in}%
    \begin{subfigure}[b]{0.166\textwidth}   
        \centering 
        {EPDN}
        \includegraphics[width=0.95\textwidth]{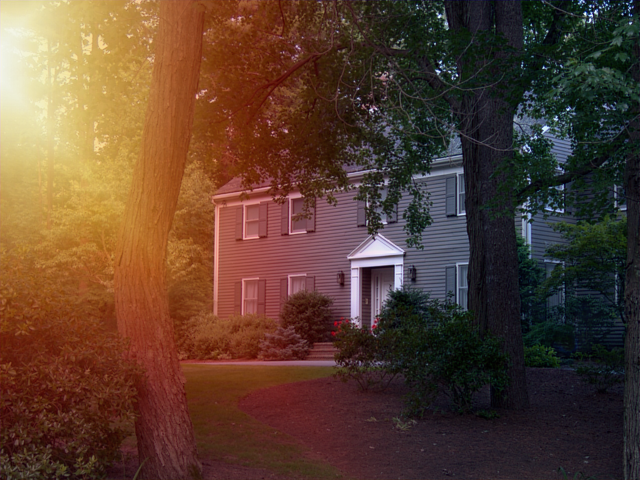}
        \label{fig:ours1}
    \end{subfigure}
    \hspace*{-0.05in}%
    \begin{subfigure}[b]{0.166\textwidth}   
        \centering 
        {Original}
        \includegraphics[width=0.95\textwidth]{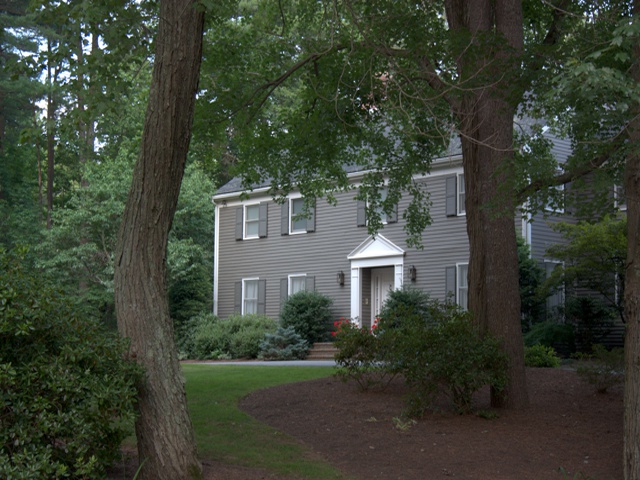}
        \label{fig:ours1}
    \end{subfigure}
    \hspace*{-0.05in}%
    \begin{subfigure}[b]{0.166\textwidth}   
        \centering 
        {Expert A}
        \includegraphics[width=0.95\textwidth]{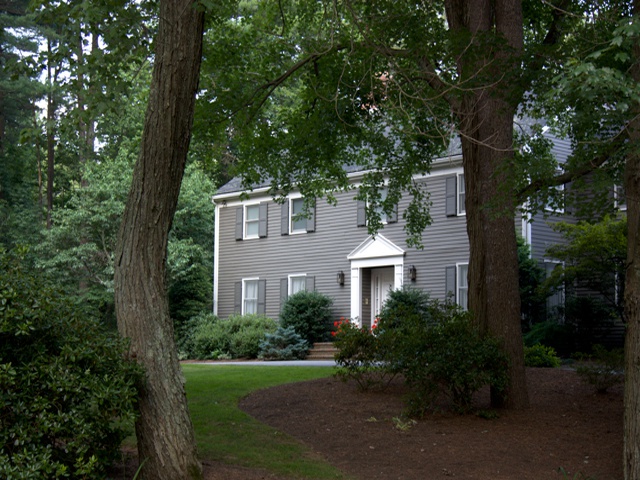}
        \label{fig:ours1}
    \end{subfigure}
        \hspace*{-0.05in}%
    \begin{subfigure}[b]{0.166\textwidth}   
        \centering 
        {Expert B}
        \includegraphics[width=0.95\textwidth]{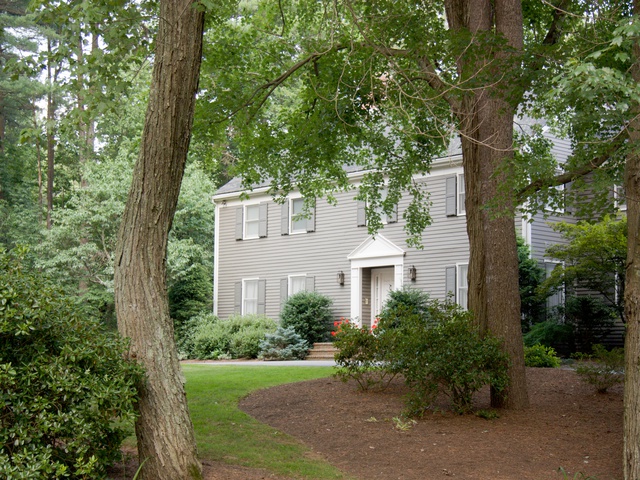}
        \label{fig:ours1}
    \end{subfigure}
    \hspace*{-0.05in}%
    \begin{subfigure}[b]{0.166\textwidth}   
        \centering 
        {Expert C}
        \includegraphics[width=0.95\textwidth]{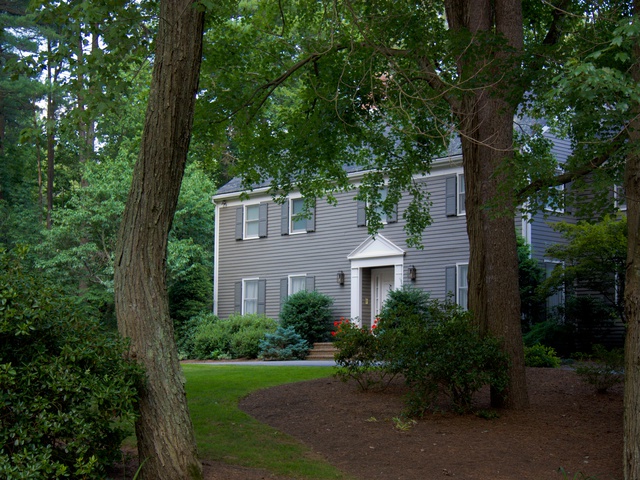}
        \label{fig:ours1}
    \end{subfigure}
        \hspace*{-0.05in}%
    \begin{subfigure}[b]{0.166\textwidth}   
        \centering 
        {Expert D}
        \includegraphics[width=0.95\textwidth]{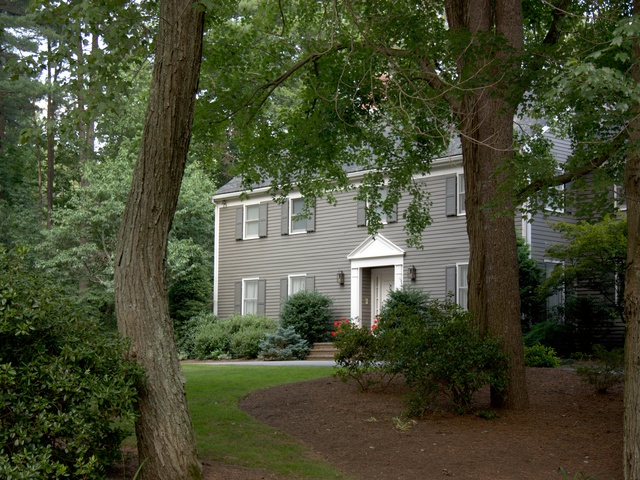}
        \label{fig:ours1}
    \end{subfigure}
    \hspace*{-0.05in}%
    \begin{subfigure}[b]{0.166\textwidth}   
        \centering 
        {Expert E}
        \includegraphics[width=0.95\textwidth]{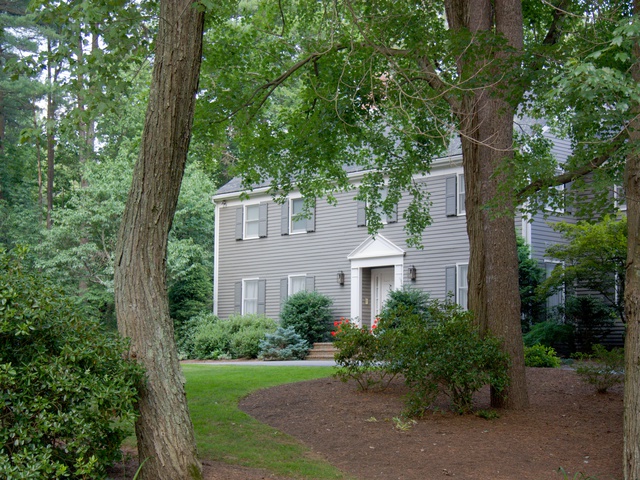}
        \label{fig:ours1}
    \end{subfigure}
    \caption{Comparison of the state-of-the-art dehazing methods on Sun-Haze dataset.}
  \label{fig:qualitative_results}
\end{figure}

\subsection{Qualitative Evaluation}

Fig.~\ref{fig:qualitative_results} depicts five hazy images from Sun-Haze dataset, and the dehazing results yielded by AOD-Net~\cite{li2017aod}, Dehaze-GLCGAN~\cite{anvari2020dehaze}, DehazeNet~\cite{cai2016dehazenet}, MSCNN~\cite{ren2016single}, and EPDN~\cite{qu2019enhanced}. In this figure we have five pairs of rows. In each pair the top row shows a hazy image followed by the results of five dehazing methods mentioned above and the second row shows the five experts' retouched images and the untouched original image to compare with.

Qualitatively, most methods were unable to remove the sunlight haze without introducing color shifting or artifacts. EPDN which is a paired image-to-image translation technique, has mainly learned to remove haze through increasing the color intensity at different channels, thus the dehazed images look visibly darker and the sunlight haze more yellow or orange than the original hazy image. EPDN also introduced artifacts to some of the generated images while partially removing haze. It also created the halo effect near edges.

Other methods also were unable to generalize to remove sunlight haze and they removed haze partially and improved the visibility to a small extent. For instance, even though AOD-Net, Dehaze-GLCGAN, Dehazenet, and MSCNN could recover the image structure similar to EPDN, they introduced color shifting and artifacts which are not visually pleasing and made the haze even more prominent. 

Dehaze-GLCGAN achieved the best/highest structural similarity index and recovered the image structure well but introduced artifacts and overexposure in particular where the sun rays lie.  

In conclusion, these methods were unable to generalize well to remove sunlight haze and performed even more poorly removing sunset hazy images which embody more varicolored haze. This questions the underlying assumptions of these methods and their practicality in the real-world scenarios. Therefore, for dehazing methods to be practical they need to be trained and tested using more realistic and practical datasets which include variety of realistic haze patterns and colors. 

%demonstrates that sunlight haze removal is a more challenging problem than traditional dehazing problem which targets removing monochromatic haze and employ synthetic datasets. In addition, dehazing methods need to be trained and tested using more realistic and practical datasets which include variety of haze patterns and colors. 

\Urlmuskip=0mu plus 1mu\relax
\bibliographystyle{acm}
\bibliography{sunhaze}

\end{document}